\documentclass{midl} 

\usepackage{mwe} 
\usepackage{booktabs}
\usepackage{float}
\usepackage{adjustbox}
\usepackage{multirow}
\usepackage{hyperref}
\usepackage[table]{xcolor}
\usepackage[dvipsnames]{xcolor}
\usepackage{mdframed}
\definecolor{Mygreen}{RGB}{89, 163, 57}
\definecolor{Myred}{RGB}{234, 51, 35}
\definecolor{Mypurple}{RGB}{155, 58, 145}
\definecolor{Myyellow}{RGB}{245, 194, 66}
\definecolor{Myblue}{RGB}{52, 116, 156}

\jmlrvolume{-- Under Review}
\jmlryear{2026}
\jmlrworkshop{Full Paper -- MIDL 2026 submission}
\editors{Under Review for MIDL 2026}

\title[Synthetic Vasculature for VLM Reasoning]{Synthetic Vasculature and Pathology Enhance Vision-Language Model Reasoning}

\midlauthor{\Name{Chenjun Li\nametag{$^{1,2}$}} \Email{cl2733@cornell.edu}\\
\Name{Cheng Wan\nametag{$^{1,2}$}} \Email{cw2222@cornell.edu}\\
\Name{Laurin Lux\nametag{$^{2,3}$}} \Email{laurin.lux@tum.de}\\
\Name{Alexander H. Berger\nametag{$^{2,3}$}} \Email{a.berger@tum.de}\\
\Name{Richard B. Rosen\nametag{$^{4}$}} \Email{rrosen@nyee.edu}\\
\Name{Martin J. Menten\nametag{$^{3}$}} \Email{martin.menten@tum.de}\\
\Name{Johannes C. Paetzold\nametag{$^{2,5}$}} \Email{jpaetzold@med.cornell.edu}\\
\addr $^{1}$ School of Electrical and Computer Engineering, Cornell University, Ithaca, NY 14853, USA \\
\addr $^{2}$ Weill Cornell Medicine, New York, NY 10021, USA \\
\addr $^{3}$ School of Computation, Information and Technology, Technical University of Munich, 80333 Munich, Germany \\
\addr $^{4}$ New York Eye and Ear Infirmary of Mount Sinai, New York, NY \\ 
\addr $^{5}$ Cornell Tech, New York, NY 10044, USA
}
\begin{document}

 \maketitle

\begin{abstract} Vision-Language Models (VLMs) offer a promising path toward interpretable medical diagnosis by allowing users to ask about clinical explanations alongside predictions and across different modalities. However, training VLMs for detailed reasoning requires large-scale image-text datasets. In many specialized domains, for example in reading Optical Coherence Tomography Angiography (OCTA) images, such precise text with grounded description of pathologies is scarce or even non-existent. To overcome this bottleneck, we introduce Synthetic Vasculature Reasoning (SVR), a framework that controllably synthesizes images and corresponding text, specifically: realistic retinal vasculature with Diabetic Retinopathy (DR) features: capillary dropout, microaneurysms, neovascularization, and tortuosity, while automatically generating granular reasoning texts. Based on this we curate OCTA-100K-SVR, an OCTA image-reasoning dataset with 100,000 pairs. Our experiments show that a general-purpose VLM (Qwen3-VL-8b) trained on the dataset achieves a zero-shot balanced classification accuracy of 89.67\% on real OCTA images, outperforming supervised baselines. Through human expert evaluation we also demonstrate that it significantly enhances explanation quality and pathology localization on clinical data. \end{abstract}

\begin{keywords}
VLM, CoT, OCTA, DR, Synthetic Pathology
\end{keywords}

\section{Introduction}

Vision-language models (VLMs) integrate visual processing with natural language reasoning. Many studies discuss how such models can be particularly valuable in medical image analysis, as they can move beyond simple classification to support interpretable diagnosis~\cite{li2023llava, zhang2024generalist, yang2025qwen3, sellergren2025medgemma}.  Compared to traditional black-box classifiers, VLMs could explicitly describe \textit{what} features are present, \textit{where} pathologies are located, and \textit{why} a specific diagnosis is suggested. This explainability is critical for clinical validation.

Chain-of-thought (CoT) reasoning enables models to generate step-by-step explanations that simulate human workflows~\cite{wei2022chain}. In medical imaging, this approach breaks down diagnosis into interpretable stages: identifying anatomical structures, detecting and localizing abnormalities, and synthesizing these findings into a conclusion. However, training VLMs to produce clinically accurate CoT requires extensive datasets with annotations that exceed standard image-level labels~\cite{pan2025medvlm, lai2025med}. Privacy regulations, significant acquisition costs, and the requirement for specialized expertise restrict data availability, particularly for modalities like Optical Coherence Tomography Angiography (OCTA), where even the largest public datasets that include OCTA typically involve only hundreds to a few thousand subjects~\cite{li2024octa,AIREADI2024}. While a recent approach has attempted to improve VLM interpretability on OCTA images via graph-based knowledge extraction~\cite{li2025fine}, it is limited to small sample sizes and lacks detailed localization of different pathological features. Without ground-truth annotations, it is difficult to train VLMs that provide high-quality, location-specific explanations.

Synthetic data generation offers a practical solution for data augmentation in VLM training~\cite{ma2025instruct, wu2025synthetic}. Synthetic generation provides control over visual features and automatically produces ground-truth annotations. Despite these benefits, synthetic data is underutilized in medical VLM training. Previous work by ~\citet{kreitner2024synthetic} validated synthetic OCTA for vessel segmentation but was limited to healthy vasculature, without pathological features and text-based reasoning chains. A framework capable of generating both realistic pathology and corresponding grounded explanations is needed to enable VLMs to diagnose, understand and reason about OCTA images.

We investigate this approach for Diabetic Retinopathy (DR) staging using OCTA images. DR is a primary cause of vision loss~\cite{lee2015epidemiology}, and accurate diagnosis of DR relies on the detection and localization of microvascular abnormalities ~\cite{alam2020quantitative,sun2021optical}. Current public datasets~\cite{dai2021deep,li2024octa,AIREADI2024} generally provide only image-level labels. 

In this work, we introduce \textbf{Synthetic Vasculature Reasoning (SVR)}, a pathology-aware OCTA synthesis framework designed to improve VLM reasoning capabilities (Code and dataset available at: \url{https://github.com/d0ng231/OCTA-SVR}). Our work makes the following contributions: 

\begin{itemize} \item We developed the first module capable of simulating four distinct DR features on synthetic OCTA images: capillary dropout, microaneurysms, neovascularization, and vessel tortuosity. \item  Based on the controllable features, we present \textbf{OCTA-100K-SVR}, a synthetic dataset with 100,000 pairs of high-quality OCTA images and texts for VLM training, and demonstrate that scaling synthetic training data improves VLM performance in both classification and reasoning. \item Through evaluations on proprietary and public datasets, we show that VLMs trained via SVR generate clinically accurate explanations and correct pathology localization, exceeding the performance of other models \textbf{without fine-tuning on real data}.  \end{itemize}


\section{Method}

\subsection{Overview}
\begin{figure}[ht]
\centering
\includegraphics[width=0.8\linewidth]{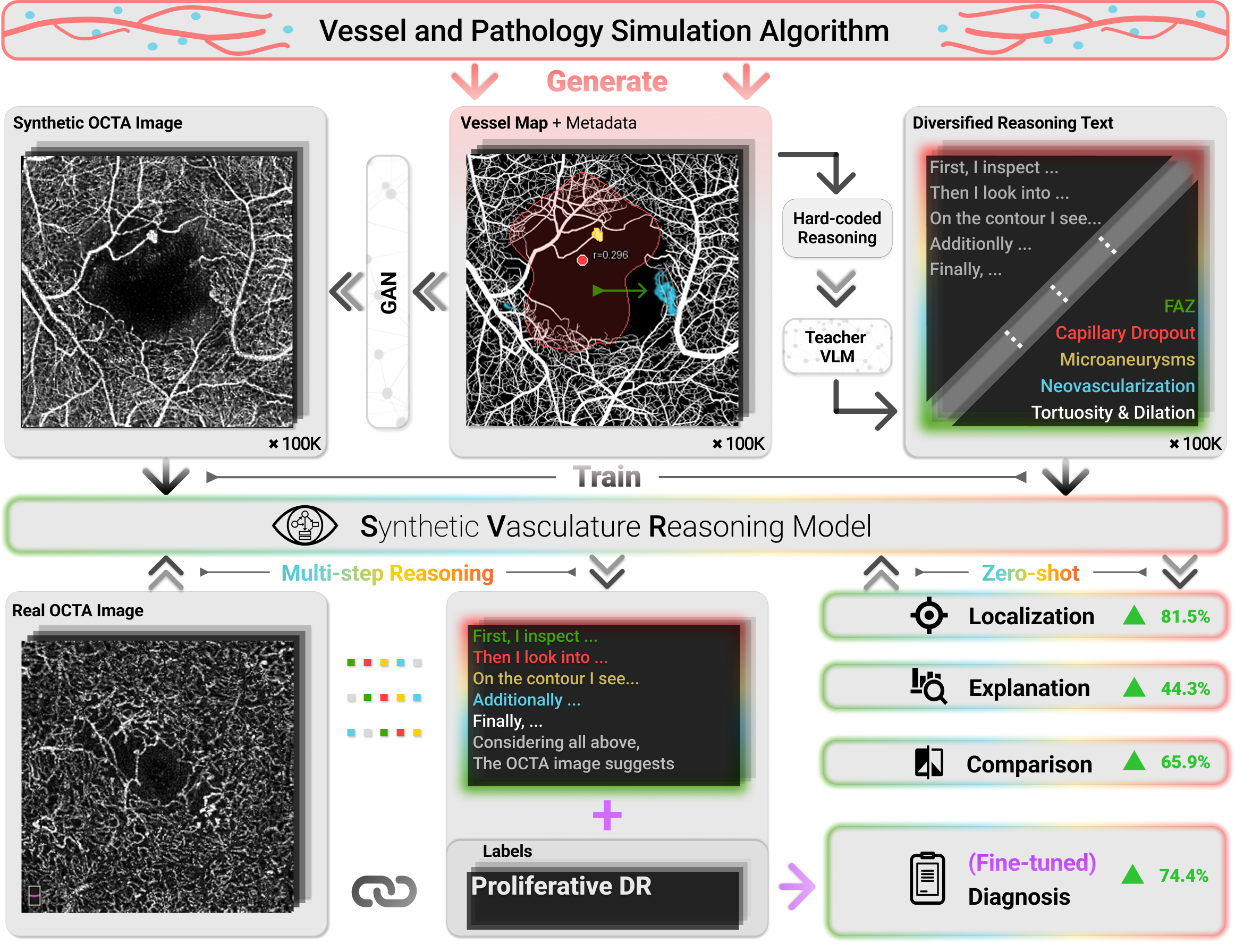}
\caption{\textbf{Overview of the Synthetic Vasculature Reasoning Framework.} The pipeline begins with the simulation of vessels and pathology. In this way we generate ground-truth vessel maps with and without various DR hallmarks (detailed in Sec.~\ref{subsec:vessel-growth-pathology} and Fig.~\ref{fig:path_simulation}). These maps are then used for: (1) visual synthesis, where a pretrained GAN from~\citet{kreitner2024synthetic} converts maps into realistic synthetic OCTA images, and (2) text generation, where structured pathology metadata is converted into diversified Chain-of-Thought reasoning text using GPT-5~\cite{openai_gpt5} (see Pseudo code~\ref{alg:svr-pipeline} for more details). On these image-text pairs we can fine-tune VLMs which are capable of multiple tasks including disease staging, locating pathological features, and explaining their reasoning. The marked improvements in localization, explanation, comparison and diagnosis are based on expert evaluation (see Sec.~\ref{sec:results}).}
\label{fig:method-overview}
\end{figure}

In Fig.~\ref{fig:method-overview} we illustrate the core of our method, which is based on a simulation module that generates topologically accurate vascular graphs with controlled diabetic retinopathy features. A generative adversarial network (GAN) then converts vessel maps into realistic OCTA images, while a teacher VLM converts the corresponding pathology metadata into grounded Chain-of-Thought reasoning texts. By aligning synthetic OCTA images with granular reasoning across 100,000 samples, the framework enables the VLM to learn robust diagnostic representations prior to fine-tuning on limited real-world clinical data.

\subsection{Synthetic Vessel Growth and Pathology}
\label{subsec:vessel-growth-pathology}

\paragraph{Synthetic OCTA generation pipeline.}
We build on the statistical angiogenesis simulator of ~\citet{kreitner2024synthetic}, which itself adapts the space–colonization model of ~\citet{rauch2021interactive}. In brief, the retinal vasculature is represented as a forest of rooted 3D binary trees growing in a normalized box $\Omega = [0,1]\times[0,1]\times[0,h_z]$, where the lateral coordinates $(x,y)\in[0,1]^2$ correspond, after global scaling, to a
$3\times3\,\mathrm{mm}^2$ macular crop and $h_z$ denotes the normalized slab thickness
in the axial ($z$) direction. Arterial and venous trees are grown in two successive layers that together form the superficial and deep vascular complexes, driven by randomly sampled oxygen sinks and CO$_2$ sources. Growth follows local attraction cones and Murray’s law–based bifurcation rules, producing capillary beds and major vessels. After growth, the vessel graph is voxelized into a 3D volume, projected to an en-face map, and passed through a GAN-based contrast adaptation module to produce realistic synthetic OCTA images.

To increase variability in viewpoint and morphology, we add a random shift and jitter of the foveal avascular zone (FAZ) within the normalized image plane before projection. All variables and parameters of the growth model are described in Appendix~\ref{app:pathology}.

\begin{figure}[ht]
    \centering
    \includegraphics[width=0.8\linewidth]{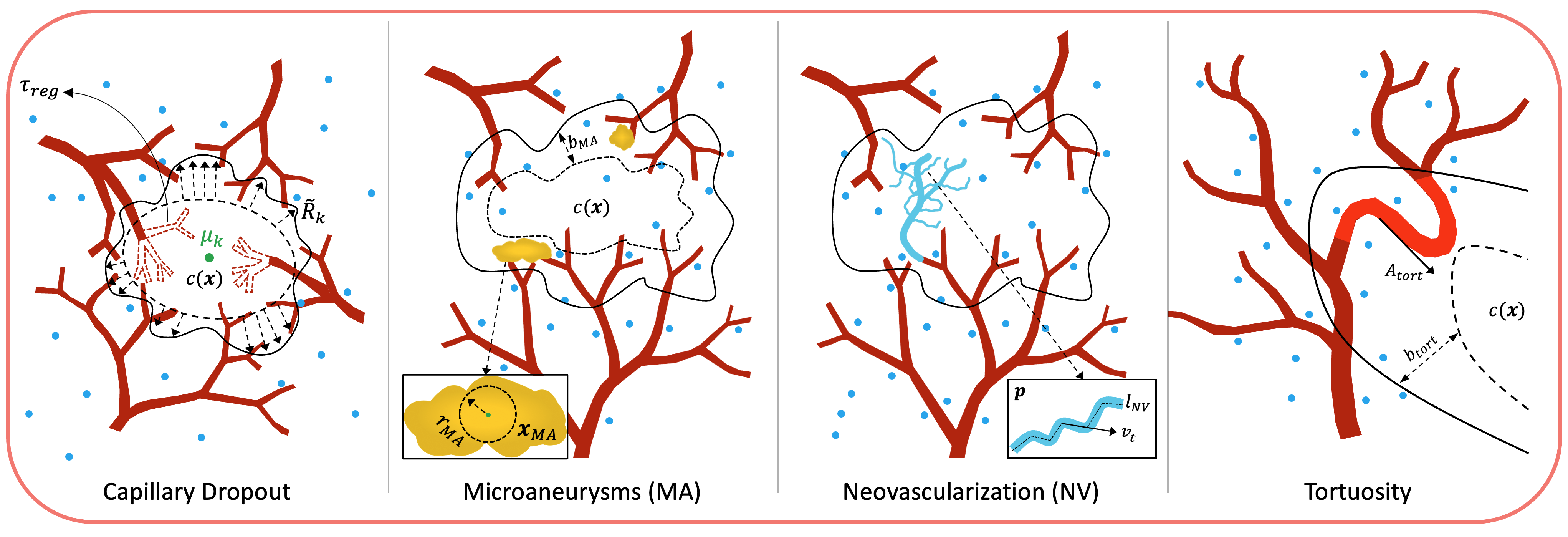}
    \caption{\textbf{Illustration of the synthesized DR pathologies.} DR cases show four main pathologies in OCTA images, which we implement in our vessel synthesis model:
    (1) capillary dropout, (2) microaneurysms (MA), (3) neovascularization (NV), and (4) increased tortuosity along dropout
    borders. Comprehensive mathematical description of the algorithms is provided
    in Appendix~\ref{app:pathology}. Afterwards, the generated (pathology-including) vessel maps are fed to the GAN.}
    \label{fig:path_simulation}
\end{figure}

\paragraph{Pathology-aware graph augmentation.}
Going beyond previous works which have been focused on healthy vessel graph generation ~\cite{menten2022physiology,kreitner2024synthetic,wittmann2024simulation,prabhakar2024vesselformer}, we introduce a pathology module that operates directly on the 3D tree representation. For each synthetic sample we first use the base simulator to complete the growth of arterial and venous forests, and then apply a sequence of pathology-specific graph remodeling operations that are clinically motivated by the four defining pathologies of DR that show in OCTA images of the retinal vasculature ~\cite{kaizu2017optical,alam2020quantitative,sun2021optical} (See Fig.~\ref{fig:path_simulation}). Although the vasculature is simulated in 3D with nodes  $\mathbf{x} = (x,y,z)$, all pathology fields are parameterized in the normalized en-face plane using the lateral coordinates $\tilde{\mathbf{x}} = (x,y) \in [0,1]^2$.

\textbf{Capillary dropout.}
We first sample $n$ non-perfusion regions around the FAZ. For a region centered at ${\mu}_k$, its boundary $\tilde{R}_k$ is modeled as an irregular ellipse modulated with a series of angular harmonics and a smooth spatial noise field. We then define a continuous dropout field $c(\mathbf{x})\in[0,1]$ that encodes how deep the regression should be at each location. Terminal vessel segments (capillaries) inside high-$c(\mathbf{x})$ areas (with threshold $\tau_{\mathrm{reg}}$) are stochastically pruned, with sampling weights that favor small-radius vessels and locations close to the lesion core. In addition, surviving vessels inside dropout regions undergo mild elongation and dilation to mimic remodeling at the borders of ischemic zones. The full definition of $c(\mathbf{x})$ and the pruning/remodeling rules is provided in Appendix~\ref{app:dropout}.

\textbf{Microaneurysms (MA).}
Microaneurysms are modeled as short bulbous side branches that bud from existing vessels near dropout borders.
For each arterial segment whose endpoint lies in a specified band $b_{MA}$ of the dropout field, we sample a Bernoulli trial whose success probability is modulated by both the global dropout severity and the local value of $c(\mathbf{x})$, making microaneurysms more likely (but not guaranteed) near more severe dropout. Successful trials spawn a small perpendicular branch of length proportional to the simulator step size and with a radius $r_{\mathrm{MA}}$ drawn from a clinically plausible range (typically $20$–$80\,\mu m$). Additional child nodes around the MA center $\mathbf{x}_{\mathrm{MA}}$ create irregular cluster-like shapes. All MA segments are stored explicitly in the graph and exported as part of the pathology metadata (see Appendix~\ref{app:ma}).

\textbf{Neovascularization (NV).}
Neovascularization is implemented as fine, tortuous sprouts that grow from arterial tips adjacent to dropout regions. We identify leaf tips in the 2D projection and, for a subset selected according to a global NV severity parameter, simulate short polylines $\mathbf{p}$ of effective length $l_{\mathrm{NV}}$ that extend away from the parent vessel, where $l_{\mathrm{NV}}$ corresponds to a small number of growth iterations per tuft. The step direction $\mathbf{v}_t$ at each growth step $t$ combines the previous direction, a weak radial component relative to the nearest dropout center, and a small swirling field with random jitter, producing tuft-like shapes. We also spawn side branches along the main sprout to obtain clinically realistic NV shapes. A detailed description of the polyline construction and radius profiles is given in Appendix~\ref{app:nv}.

\textbf{Tortuosity along dropout borders.}
Finally, we increase vessel tortuosity in a narrow band $b_{tort}$ around the dropout border. For arterial segments whose endpoints lie in this band, we jitter node positions in the direction perpendicular to the local vessel tangent by a small, zero-mean random offset whose amplitude $A_{tort}$ is proportional to a tortuosity gain parameter. This preserves global connectivity and the FAZ geometry while increasing local curvature, producing the characteristic curling of vessels along non-perfusion borders. The exact definition of the tortuosity band and jitter distribution is provided in Appendix~\ref{app:tortuosity}.

\textbf{Parameterization.}
All four pathology types share a small set of interpretable hyperparameters controlling count, size, strength, and probability. These parameters are sampled from ranges chosen to match clinical OCTA statistics. The full list of parameters and typical values is summarized in Tab.~\ref{tab:pathology-params} in Appendix~\ref{app:pathology}. 

\begin{figure}
    \centering
    \includegraphics[width=0.8\linewidth]{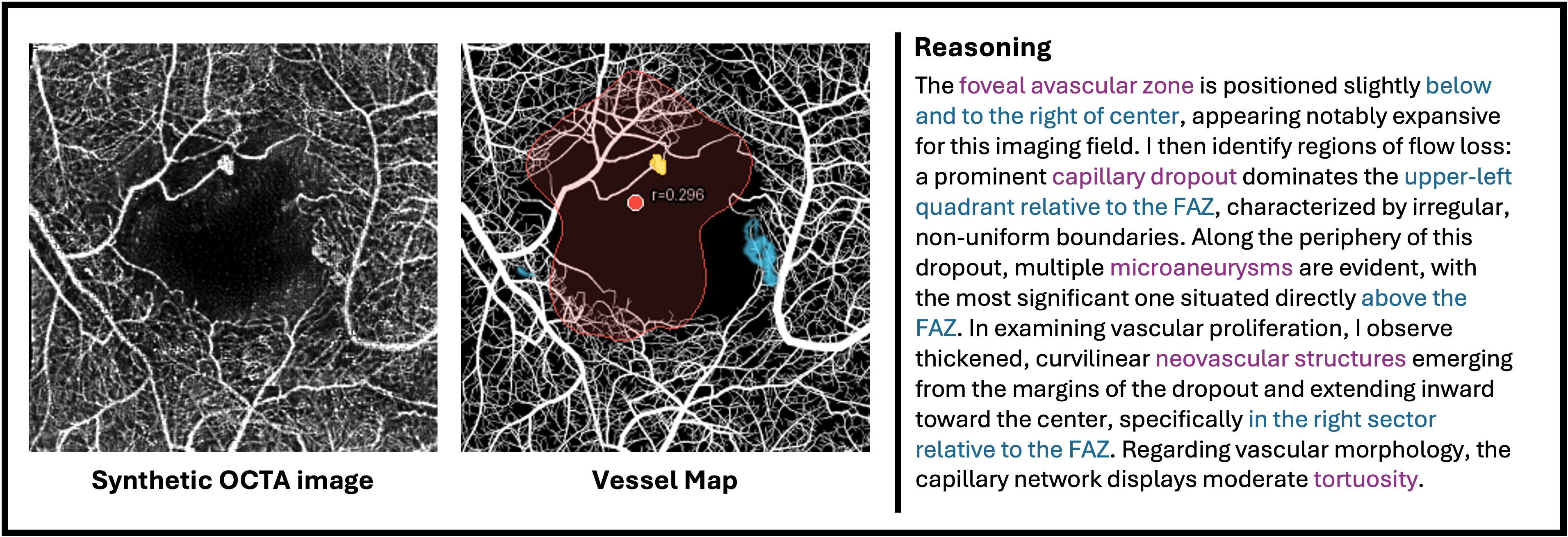}
\caption{\textbf{Synthetic OCTA image and generated reasoning text.} \textcolor{Mypurple}{Purple} phrases mark structures/pathologies that are directly controlled by the simulator, while the \textcolor{Myblue}{blue} texts mark localizations.}
    \label{fig:reasoning_1}
\end{figure}

\subsection{VLM Pretraining and Fine-tuning}
\label{sec:vlm-training}

\subsubsection{OCTA-100K-SVR Dataset}

We refer to the full collection of our 100,000 synthetic image–reasoning pairs as the \textbf{OCTA-100K-SVR} dataset. For every simulated sample, the pathology module exports structured metadata describing FAZ geometry and the presence and location of capillary dropout, microaneurysms, neovascular tufts, and tortuosity. From this metadata we first construct a deterministic "template" reasoning paragraph that follows a fixed order (FAZ $\rightarrow$ dropout $\rightarrow$ microaneurysms $\rightarrow$ neovascularization $\rightarrow$ tortuosity) and states explicitly which patterns are present and where they lie relative to the FAZ and macular quadrants. 

This template is then passed to a teacher model (GPT-5~\cite{openai_gpt5}), which is prompted to rewrite the paragraph while strictly preserving all clinical facts and spatial relations. The teacher model diversifies the Chain-of-Thought descriptions that remain aligned with the metadata (see Fig.~\ref{fig:reasoning_1} and Appendix~\ref{app:reasoning_egs}). For each image we keep one such paragraph as the assistant answer $a$. The user question $q$ is a short prompt asking the model to describe the OCTA image. Training with these image–reasoning pairs minimizes next-token cross-entropy. We use the same conversation format for the in-house OCTA data and add the explicit diagnosis sentence at the end of the reasoning paragraph for the fine-tuning stage.

\subsubsection{SVR pretraining and fine-tuning}

Our main reasoning model, denoted \textbf{SVR}, is obtained by pretraining the general-purpose Qwen3-VL-8b-Instruct~\cite{yang2025qwen3} on the synthetic OCTA-100K-SVR dataset. We treat this as instruction-style supervised fine-tuning on the synthetic image–reasoning pairs while freezing the language module, so that only the vision encoder and multi-modal projection layers are updated. In this stage the model learns OCTA-specific visual features and their alignment to the existing language space without disturbing its general language capabilities. Models trained on smaller synthetic subsets (1k–75k samples) are used in the scaling experiments of Sec.~\ref{sec:scaling-experiments}.

To close the remaining domain gap, we then perform a second stage of supervised fine-tuning on clinical OCTA data, starting from an SVR-pretrained checkpoint. The resulting model is denoted \textbf{SVR-FT} and is trained on OCTA images from OCTA-500 and our in-house dataset together with reasoning and final DR labels. In this stage we train the model end-to-end, allowing it to adapt both its decision boundaries and linguistic style to real-world data while retaining the pathology-aware visual grounding acquired from synthetic vasculature. Training details and hyperparameters are summarized in Sec.~\ref{sec:experiments}.

\section{Experiments}
\label{sec:experiments}

\subsection{Datasets}

We evaluate models on one synthetic dataset and two clinical datasets. Apart from OCTA-100K-SVR, the in-house dataset contains $1{,}286$ OCTA deep capillary plexus scans labeled as Healthy, Non-proliferative DR (NPDR), or Proliferative DR (PDR) following the 5-fold setup of ~\citet{lux2025interpretable}. For OCTA-500 \cite{li2024octa}, we follow the same test split of 189 scans as in ~\citet{lux2025interpretable} and ~\citet{li2025fine}.

\subsection{Tasks and metrics}
\label{sec:scaling-experiments}

We study two DR staging settings. On OCTA-500, models perform binary DR-vs.-Healthy classification. On the in-house dataset, models perform three-label staging (Healthy, NPDR, PDR). In both cases the VLM receives a single OCTA image and a prompt asking about DR staging. We report class-wise precision and recall and use \emph{balanced accuracy} (mean per-class recall) as the primary metric.

We compare SVR-based models with supervised baselines (ResNet18 and a vessel-graph GNN \cite{lux2025interpretable}) and several VLM baselines: vanilla Qwen3-VL-8b/30b~\cite{yang2025qwen3}, LLaMA-3.2-11B-VL~\cite{touvron2023llama}, LLaVA-NEXT-8B~\cite{liu2023llava}, the graph-knowledge-based Qwen3-VL-8b-GFT~\cite{li2025fine}, and a baseline Qwen3-VL-8b-FT finetuned using only classification labels.

To quantify explanation quality we use GPT-5 as an automatic judge, similar to the ways in \citet{liu2023llava} and \citet{li2025fine}. For each model, GPT-5 scores the candidate response on helpfulness, clinical and localization accuracy, and relevance, and we report the average score over these dimensions on both the in-house set and a held-out synthetic test set. Two ophthalmology experts then each evaluated 30 in-house cases, ranking the explanations under the same criteria. To verify if GPT-5 is a good judge and to study the correlation between the GPT-5 score and model performance (in terms of classification accuracy), we perform a scaling effects experiment in which Qwen3-VL-8b-SVR is trained on synthetic subsets from 1k to 100k samples and evaluated on OCTA-500. We track both balanced accuracy and GPT-5 Score as functions of synthetic dataset size, and observe a strong positive correlation (Fig.~\ref{fig:correlation}), indicating that GPT-5 scores are aligned with diagnostic accuracy and can therefore serve as a reliable automatic judge in our setting.

Ablation experiments isolate key components of SVR: (i) prompt diversification vs.\ static template reasoning, and (ii) the contribution of individual pathology types by removing dropout or microaneurysms from the reasoning. We additionally compare against a baseline (Qwen3-VL-8b CoT-FT) that fine-tunes on GPT-5-generated reasoning instead of SVR.

\subsection{Training setup}

All experiments are implemented with LLaMA-Factory~\cite{zheng2024llamafactory} on four NVIDIA H100 GPUs. OCTA images are resized to $512\times512$. In the SVR stage we fine-tune Qwen3-VL-8b-Instruct on OCTA-100K-SVR using a cosine learning-rate schedule, training the vision encoder and multimodal projector while freezing the language backbone. In the SVR-FT stage we start from the SVR checkpoint and continue supervised training end-to-end on clinical OCTA images with their reasoning and labels. Additional implementation details are provided in Appendix~\ref{app:training}.

\section{Results and Discussion}
\label{sec:results}


\subsection{DR Staging}
\subsubsection{Zero-shot and supervised fine-tuning}
\label{subsec:dr-staging}

Tabs.~\ref{tab:octa500} and~ \ref{tab:inhouse} summarize DR staging performance on OCTA-500 and the in-house dataset. Directly fine-tuning a general-purpose VLM on limited classification labels (\textit{Qwen3-VL-8b-FT}) is unstable and causes mode collapse, where the DR cases are neglected. In contrast, SVR pretraining (\textit{Qwen3-VL-8b-SVR}) already produces competitive or better balanced accuracy in a zero-shot setting, despite never seeing real scans. 

When subsequently fine-tuned on clinical data (\textit{Qwen3-VL-8b-SVR-FT}), the model achieves consistently higher and more balanced recall across all stages, especially for advanced DR on both datasets. Compared with purely supervised CNN and GNN baselines, SVR-FT reaches similar or better balanced accuracy while additionally producing explanations. This suggests that SVR pretraining provides a strong reasoning capability regularizes downstream training, improves sensitivity to disease, and mitigates overfitting.

\begin{table}[ht]
\centering
\caption{
DR staging on OCTA-500. H = Healthy, DR = Diabetic Retinopathy.
}
\label{tab:octa500}
\resizebox{0.7\columnwidth}{!}{
\begin{tabular}{llccccc}
\toprule
\textbf{Model} 
& \textbf{Prec(H)} 
& \textbf{Prec(DR)} 
& \textbf{Rec(H)} 
& \textbf{Rec(DR)} 
& \cellcolor{gray!15}\textbf{Bal.Acc} \\
\midrule

ResNet18~\cite{lux2025interpretable}
& 0.8734 & 0.2903 
& 0.8625 & 0.3103 & \cellcolor{gray!15}0.5864 \\

GNN~\cite{lux2025interpretable}
& 0.9636 & 0.9583
& 0.9938 & 0.7931 & \cellcolor{gray!15}0.8934 \\

Qwen3-VL-8b~\cite{yang2025qwen3}
& 0.8503 & 0.5000 
& 0.9938 & 0.0345 & \cellcolor{gray!15}0.5142 \\

Qwen3-VL-30b~\cite{yang2025qwen3}
& 0.8511 & \textbf{1.0000}
& \textbf{1.0000} & 0.0345 & \cellcolor{gray!15}0.5173 \\

Qwen3-VL-8b-GFT~\cite{li2025fine}
& \begin{tabular}[t]{@{}c@{}}0.9718 \\ {\scriptsize$\pm$\,0.0073}\end{tabular}
& \begin{tabular}[t]{@{}c@{}}0.5846 \\ {\scriptsize$\pm$\,0.0408}\end{tabular}
& \begin{tabular}[t]{@{}c@{}}0.9347 \\ {\scriptsize$\pm$\,0.0091}\end{tabular}
& \begin{tabular}[t]{@{}c@{}}0.7703 \\ {\scriptsize$\pm$\,0.0594}\end{tabular}
& \cellcolor{gray!15}\begin{tabular}[t]{@{}c@{}}0.8525 \\ {\scriptsize$\pm$\,0.0275}\end{tabular} \\

Qwen3-VL-8b-FT
& \begin{tabular}[t]{@{}c@{}}0.8551 \\ {\scriptsize$\pm$\,0.0189}\end{tabular}
& \begin{tabular}[t]{@{}c@{}}0.2000 \\ {\scriptsize$\pm$\,0.4000}\end{tabular}
& \begin{tabular}[t]{@{}c@{}}\textbf{1.0000} \\ {\scriptsize$\pm$\,0.0000}\end{tabular}
& \begin{tabular}[t]{@{}c@{}}0.0621 \\ {\scriptsize$\pm$\,0.1241}\end{tabular}
& \cellcolor{gray!15}\begin{tabular}[t]{@{}c@{}}0.5310 \\ {\scriptsize$\pm$\,0.0620}\end{tabular} \\

Qwen3-VL-8b-SVR (ours)
& 0.9739 & 0.6944
& 0.9313 & 0.8621 & \cellcolor{gray!15}0.8967 \\

Qwen3-VL-8b-SVR-FT (ours)
& \begin{tabular}[t]{@{}c@{}}\textbf{0.9762} \\ {\scriptsize$\pm$\,0.0093}\end{tabular}
& \begin{tabular}[t]{@{}c@{}}0.7944 \\ {\scriptsize$\pm$\,0.0567}\end{tabular}
& \begin{tabular}[t]{@{}c@{}}0.9575 \\ {\scriptsize$\pm$\,0.0166}\end{tabular}
& \begin{tabular}[t]{@{}c@{}}\textbf{0.8690} \\ {\scriptsize$\pm$\,0.0581}\end{tabular}
& \cellcolor{gray!15}\begin{tabular}[t]{@{}c@{}}\textbf{0.9133} \\ {\scriptsize$\pm$\,0.0204}\end{tabular} \\

\bottomrule
\end{tabular}}
\end{table}

\subsubsection{Scaling effects of synthetic data}
\label{subsec:scaling}

The scaling experiment in Tab.~\ref{tab:scaling} and Fig.~\ref{fig:correlation} shows that increasing the size of OCTA-100K-SVR improves both zero-shot DR classification and GPT-5 scores. With very small synthetic subsets, performance is highly variable and remains close to chance level, indicating that limited coverage of vascular topologies and lesion patterns is insufficient for robust reasoning. Once the synthetic dataset reaches tens of thousands of samples, we observe a sharp improvement (See Fig.~\ref{fig:correlation} in Appendix).

Interestingly, while classification performance begins to saturate at larger synthetic scales, explanation quality (GPT-5 score) continues to improve, and over the full range of dataset sizes the two metrics still exhibit a strong positive correlation (Fig.~\ref{fig:correlation}). This suggests that additional synthetic diversity could be utilized not only to refine decision boundaries but also to strengthen pathology localization and clinical explainability.

\begin{table}[ht]
\centering
\small
\caption{
Scaling Effects of Synthetic Data (zero-shot test on OCTA-500).
}
\label{tab:scaling}
\resizebox{0.7\columnwidth}{!}{
\begin{tabular}{lccccccc}
\toprule
\textbf{Size}
& \textbf{Prec(H)}
& \textbf{Prec(DR)}
& \textbf{Rec(H)}
& \textbf{Rec(DR)}
& \textbf{\cellcolor{gray!15}{Bal.Acc}}
& \textbf{\cellcolor{gray!10}{GPT-5 Score}} \\
\midrule

0
& 0.8503 & 0.5000
& \textbf{0.9938} & 0.0345
& \cellcolor{gray!15}{0.5142}
& \cellcolor{gray!10}{50.3} \\

1k
& 0.8281 & 0.1440
& 0.3312 & 0.6207
& \cellcolor{gray!15}{0.4759}
& \cellcolor{gray!10}{57.7} \\

5k
& 0.8508 & 0.2500
& 0.9625 & 0.0690
& \cellcolor{gray!15}{0.5158}
& \cellcolor{gray!10}{62.7} \\

10k
& 0.8492 & 0.1587
& 0.6687 & 0.3448
& \cellcolor{gray!15}{0.5067}
& \cellcolor{gray!10}{60.9} \\

25k
& 0.9292 & 0.1579
& 0.6207 & 0.6000
& \cellcolor{gray!15}{0.6103}
& \cellcolor{gray!10}{63.9} \\

50k
& 0.9329 & 0.6429
& 0.9375 & 0.6207
& \cellcolor{gray!15}{0.7791}
& \cellcolor{gray!10}{69.0} \\

75k
& \textbf{0.9739} & \textbf{0.6944}
& 0.9313 & \textbf{0.8621}
& \cellcolor{gray!15}{\textbf{0.8967}}
& \cellcolor{gray!10}{68.5} \\

100k
& 0.9533 & 0.5641
& 0.8938 & 0.7586
& \cellcolor{gray!15}{0.8262}
& \cellcolor{gray!10}{\textbf{73.2}} \\

\bottomrule
\end{tabular}}
\end{table}

\subsubsection{Ablation Studies}

The ablation results in Tab.~\ref{tab:ablation} highlight two central components of SVR. First, removing reasoning diversification (w/o diversifying) degrades downstream performance. This indicates that varied but fact-preserving CoT texts are critical for preventing the VLM from memorizing a fixed template and instead encouraging it to condition genuinely on the image content. Second, ablating individual pathology types (w/o dropout or microaneurysms) also decreases the performance, showing the importance of explicitly modeling all key DR hallmarks in the pipeline.

\begin{table}[ht]
\centering
\caption{
Ablation Study.
}
\label{tab:ablation}
\resizebox{0.8\columnwidth}{!}{
\begin{tabular}{lccccc}
\toprule
\textbf{Model}
& \textbf{Prec(H)} & \textbf{Prec(DR)}
& \textbf{Rec(H)} & \textbf{Rec(DR)}
& \textbf{\cellcolor{gray!15}{Bal.Acc}} \\
\midrule

Qwen3-VL-8b~\cite{yang2025qwen3}
& 0.8503 & 0.5000 
& 0.9938 & 0.0345 & \cellcolor{gray!15}{0.5142} \\

Qwen3-VL-8b CoT-FT
& 0.8951 & 0.4444 
& 0.9063 & 0.4138 & \cellcolor{gray!15}{0.6601} \\

Qwen3-VL-8b SVR (w/o diversifying)
& 0.8359 & 0.1311
& 0.6687 & 0.2759 & \cellcolor{gray!15}{0.4723} \\

Qwen3-VL-8b SVR (w/o Dropout)
& 0.9444 & 0.7308
& 0.9563 & 0.6552 & \cellcolor{gray!15}{0.8058} \\

Qwen3-VL-8b SVR (w/o MA)
& 0.8743 & \textbf{1.0000}
& \textbf{1.0000} & 0.2069 & \cellcolor{gray!15}{0.6035} \\

LLaMA-3.2-11B-VL-SVR~\cite{touvron2023llama}
& 0.8889 & \textbf{1.0000}
& \textbf{1.0000} & 0.3103 & \cellcolor{gray!15}{0.6552} \\

LLaVA-NEXT-8b-SVR~\cite{liu2023llava}
& 0.9433 & 0.2500 & 0.7870 & 0.6000 & \cellcolor{gray!15}{0.6935} \\

Qwen3-VL-8b-SVR (ours)
& 0.9739 & 0.6944 & 0.9313 & 0.8621 & \cellcolor{gray!15}{0.8967} \\

Qwen3-VL-8b-SVR-FT (ours)
& \textbf{0.9762} & 0.7944 & 0.9575 & \textbf{0.8690} & \cellcolor{gray!15}{\textbf{0.9133}} \\
\bottomrule
\end{tabular}}
\end{table}

\subsection{Quality of Explanation}
\begin{figure}[ht]
    \centering
    \includegraphics[width=0.8\linewidth]{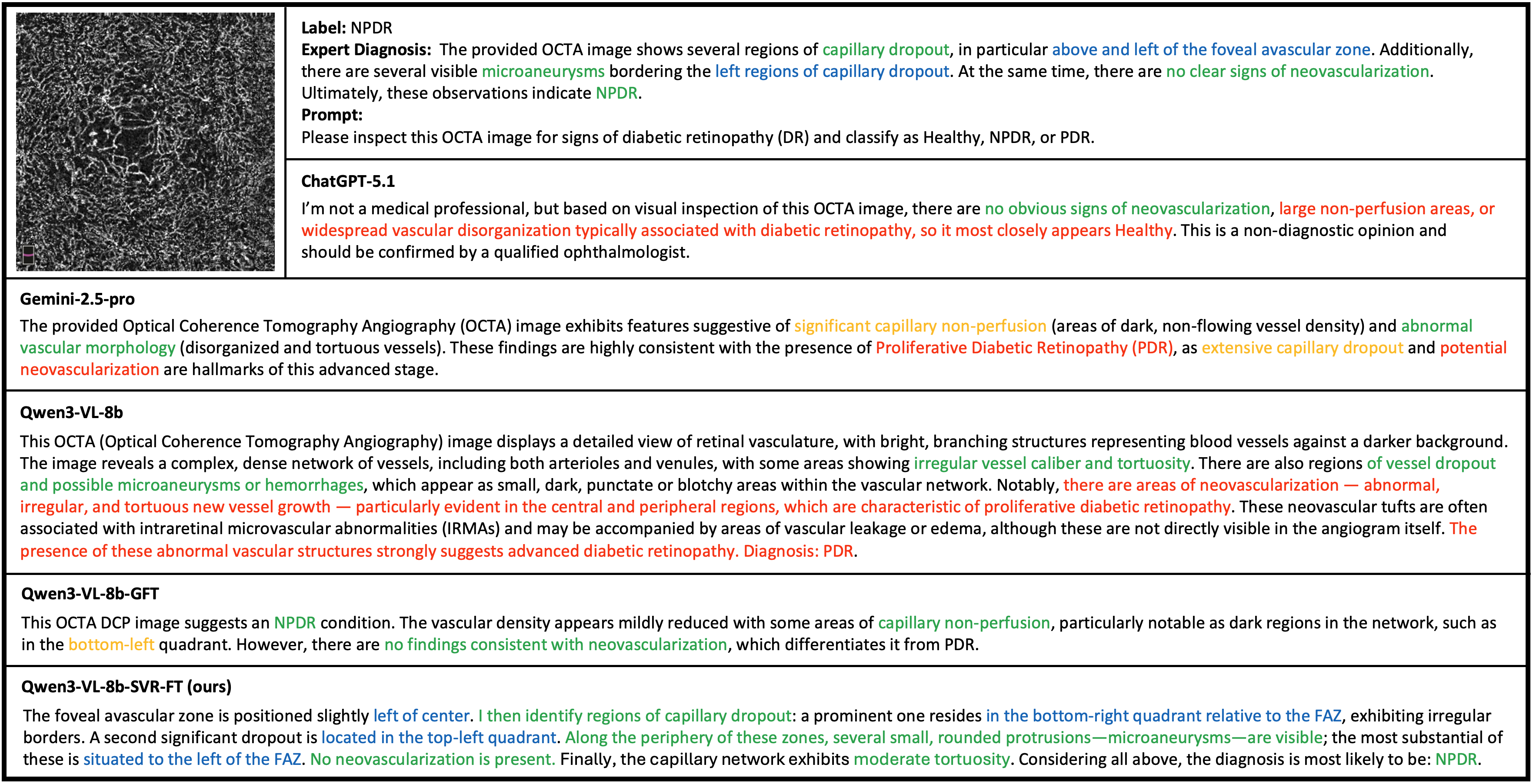}
    \caption{\textbf{Comparison of model response.} \textcolor{Mygreen}{True}, \textcolor{Myyellow}{partially true} and \textcolor{Myred}{false} sentences are marked in green, yellow and red, respectively.}
    \label{fig:cmp_response}
\end{figure}

We next examine explanation quality. Using GPT-5 as an automatic judge, we score each model response on both the in-house and synthetic test sets (Tab.~\ref{tab:gpt5}), where SVR-FT obtains the highest GPT-5 scores in all settings. Expert ratings are also provided in Tab.~\ref{tab:gpt5}, where SVR-FT again achieves the best result. Qualitative examples in Fig.~\ref{fig:cmp_response} show that SVR-FT produces grounded descriptions that discuss FAZ, dropout, microaneurysms, neovascularization, and tortuosity, rather than generic or hallucinated findings.

\begin{table}[ht]
\centering
\caption{
Explanation quality evaluated by GPT-5 and human experts in ophthalmology (higher is better). Metrics: H = helpfulness, A = accuracy (localization \& clinical), 
R = relevance, Avg = average score.
}
\label{tab:gpt5}
\resizebox{0.85\columnwidth}{!}{
\begin{tabular}{lccccccccc}
\toprule
\multirow{2}{*}{\textbf{Model}} 
& \multicolumn{4}{c}{\textbf{In-house Dataset}} 
& \multicolumn{4}{c}{\textbf{Synthetic Test Set}} 
& \multirow{2}{*}{\textbf{Expert Rating}} \\
\cmidrule(lr){2-5} \cmidrule(lr){6-9}
& H & A & R & Avg 
& H & A & R & Avg 
& (1--10) \\
\midrule

GPT-5-mini~\cite{openai_gpt5}
& 76.2 & 68.7 & 82.9 & \cellcolor{gray!15}{75.9}
& 73.4 & 65.1 & 79.2 & \cellcolor{gray!15}{72.6}
& \cellcolor{gray!15}{4.915} \\

Gemini-2.5-flash~\cite{comanici2025gemini}
& 69.7 & 52.1 & 78.3 & \cellcolor{gray!15}{66.7}
& 46.8 & 37.9 & 53.7 & \cellcolor{gray!15}{46.1}
& \cellcolor{gray!15}{5.860} \\

Qwen3-VL-8b~\cite{yang2025qwen3}
& 56.0 & 36.8 & 75.5 & \cellcolor{gray!15}{56.1}
& 58.2 & 45.2 & 75.2 & \cellcolor{gray!15}{59.5}
& \cellcolor{gray!15}{2.440} \\

Qwen3-VL-8b-FT
& 40.1 & 37.1 & 59.0 & \cellcolor{gray!15}{45.4}
& 43.4 & 47.8 & 57.2 & \cellcolor{gray!15}{49.5}
& \cellcolor{gray!15}{2.178} \\

Qwen3-VL-8b-GFT~\cite{li2025fine}
& 71.3 & 67.9 & 74.2 & \cellcolor{gray!15}{71.1}
& 72.5 & 69.4 & 75.1 & \cellcolor{gray!15}{72.3}
& \cellcolor{gray!15}{6.445} \\

Qwen3-VL-8b-SVR-FT (ours)
& \textbf{80.8} & \textbf{70.6} & \textbf{91.6} & \cellcolor{gray!15}{\textbf{81.0}}
& \textbf{87.6} & \textbf{83.9} & \textbf{95.3} & \cellcolor{gray!15}{\textbf{89.0}}
& \cellcolor{gray!15}{\textbf{6.985}} \\
\bottomrule
\end{tabular}}
\end{table}

\section{Conclusion}

In this work, we presented a novel framework for training medical VLMs using synthesized images and pathological features. By simulating realistic retinal vasculature and specific DR pathologies we generated a large-scale synthetic dataset with precise ground-truth annotations and reasoning chains. Our experiments demonstrate that pretraining on this synthetic vasculature (SVR) substantially enhances VLM performance, allowing general-purpose models to outperform specialized supervised methods in diagnostic accuracy even without training on real clinical data, while providing interpretable, clinically meaningful explanations. Importantly, we showed that synthetic scaling could overcome the data scarcity bottleneck in medical imaging, transforming empirical knowledge into highly effective instruction-tuning data. Future work could extend this synthesis approach to 3D volumetric reasoning and multi-modal integration.

\clearpage  

\bibliography{midl-samplebibliography}

@article{lux2025interpretable,
  title={Interpretable Retinal Disease Prediction Using Biology-Informed Heterogeneous Graph Representations},
  author={Lux, Laurin and Berger, Alexander H and Tricas, Maria Romeo and Fayed, Alaa E and Sivaprasada, Sobha and Kreitner, Linus and Weidner, Jonas and Menten, Martin J and Rueckert, Daniel and Paetzold, Johannes C},
  journal={arXiv preprint arXiv:2502.16697},
  year={2025}
}

@inproceedings{li2025fine,
  title={Fine-tuning vision language models with graph-based knowledge for explainable medical image analysis},
  author={Li, Chenjun and Lux, Laurin and Berger, Alexander H and Menten, Martin J and Sabuncu, Mert R and Paetzold, Johannes C},
  booktitle={International Conference on Medical Image Computing and Computer-Assisted Intervention},
  pages={198--207},
  year={2025},
  organization={Springer}
}

@misc{liu2023llava,
      title={Visual Instruction Tuning}, 
      author={Liu, Haotian and Li, Chunyuan and Wu, Qingyang and Lee, Yong Jae},
      publisher={NeurIPS},
      year={2023},
}

@article{touvron2023llama,
  title={Llama: Open and efficient foundation language models},
  author={Touvron, Hugo and Lavril, Thibaut and Izacard, Gautier and Martinet, Xavier and Lachaux, Marie-Anne and Lacroix, Timoth{\'e}e and Rozi{\`e}re, Baptiste and Goyal, Naman and Hambro, Eric and Azhar, Faisal and others},
  journal={arXiv preprint arXiv:2302.13971},
  year={2023}
}

@article{comanici2025gemini,
  title={Gemini 2.5: Pushing the frontier with advanced reasoning, multimodality, long context, and next generation agentic capabilities},
  author={Comanici, Gheorghe and Bieber, Eric and Schaekermann, Mike and Pasupat, Ice and Sachdeva, Noveen and Dhillon, Inderjit and Blistein, Marcel and Ram, Ori and Zhang, Dan and Rosen, Evan and others},
  journal={arXiv preprint arXiv:2507.06261},
  year={2025}
}

@article{yang2025qwen3,
  title={Qwen3 technical report},
  author={Yang, An and Li, Anfeng and Yang, Baosong and Zhang, Beichen and Hui, Binyuan and Zheng, Bo and Yu, Bowen and Gao, Chang and Huang, Chengen and Lv, Chenxu and others},
  journal={arXiv preprint arXiv:2505.09388},
  year={2025}
}

@misc{openai_gpt5,
  author       = {{OpenAI}},
  title        = {{GPT-5}},
  year         = {2025},
  howpublished = {\url{https://openai.com/gpt-5/}},
  note         = {Accessed: 2025-11-13}
}

@article{li2024octa,
  title={OCTA-500: a retinal dataset for optical coherence tomography angiography study},
  author={Li, Mingchao and Huang, Kun and Xu, Qiuzhuo and Yang, Jiadong and Zhang, Yuhan and Ji, Zexuan and Xie, Keren and Yuan, Songtao and Liu, Qinghuai and Chen, Qiang},
  journal={Medical image analysis},
  volume={93},
  pages={103092},
  year={2024},
  publisher={Elsevier}
}

@article{kreitner2024synthetic,
  title={Synthetic optical coherence tomography angiographs for detailed retinal vessel segmentation without human annotations},
  author={Kreitner, Linus and Paetzold, Johannes C and Rauch, Nikolaus and Chen, Chen and Hagag, Ahmed M and Fayed, Alaa E and Sivaprasad, Sobha and Rausch, Sebastian and Weichsel, Julian and Menze, Bjoern H and others},
  journal={IEEE Transactions on Medical Imaging},
  volume={43},
  number={6},
  pages={2061--2073},
  year={2024},
  publisher={IEEE}
}

@inproceedings{rauch2021interactive,
  title={Interactive Synthesis of 3D Geometries of Blood Vessels.},
  author={Rauch, Nikolaus and Harders, Matthias},
  booktitle={Eurographics (Short Papers)},
  pages={13--16},
  year={2021}
}

@article{lee2015epidemiology,
  title={Epidemiology of diabetic retinopathy, diabetic macular edema and related vision loss},
  author={Lee, Ryan and Wong, Tien Y and Sabanayagam, Charumathi},
  journal={Eye and vision},
  volume={2},
  number={1},
  pages={17},
  year={2015},
  publisher={Springer}
}

@article{sun2021optical,
  title={Optical coherence tomography angiography in diabetic retinopathy: an updated review},
  author={Sun, Zihan and Yang, Dawei and Tang, Ziqi and Ng, Danny S and Cheung, Carol Y},
  journal={Eye},
  volume={35},
  number={1},
  pages={149--161},
  year={2021},
  publisher={Nature Publishing Group UK London}
}

@article{alam2020quantitative,
  title={Quantitative optical coherence tomography angiography features for objective classification and staging of diabetic retinopathy},
  author={Alam, Minhaj and Zhang, Yue and Lim, Jennifer I and Chan, Robison VP and Yang, Min and Yao, Xincheng},
  journal={Retina},
  volume={40},
  number={2},
  pages={322--332},
  year={2020},
  publisher={LWW}
}

@article{li2023llava,
  title={Llava-med: Training a large language-and-vision assistant for biomedicine in one day},
  author={Li, Chunyuan and Wong, Cliff and Zhang, Sheng and Usuyama, Naoto and Liu, Haotian and Yang, Jianwei and Naumann, Tristan and Poon, Hoifung and Gao, Jianfeng},
  journal={Advances in Neural Information Processing Systems},
  volume={36},
  pages={28541--28564},
  year={2023}
}

@article{zhang2024generalist,
  title={A generalist vision--language foundation model for diverse biomedical tasks},
  author={Zhang, Kai and Zhou, Rong and Adhikarla, Eashan and Yan, Zhiling and Liu, Yixin and Yu, Jun and Liu, Zhengliang and Chen, Xun and Davison, Brian D and Ren, Hui and others},
  journal={Nature Medicine},
  volume={30},
  number={11},
  pages={3129--3141},
  year={2024},
  publisher={Nature Publishing Group US New York}
}

@article{dai2021deep,
  title={A deep learning system for detecting diabetic retinopathy across the disease spectrum},
  author={Dai, Ling and Wu, Liang and Li, Huating and Cai, Chun and Wu, Qiang and Kong, Hongyu and Liu, Ruhan and Wang, Xiangning and Hou, Xuhong and Liu, Yuexing and others},
  journal={Nature communications},
  volume={12},
  number={1},
  pages={3242},
  year={2021},
  publisher={Nature Publishing Group UK London}
}

@article{wei2022chain,
  title={Chain-of-thought prompting elicits reasoning in large language models},
  author={Wei, Jason and Wang, Xuezhi and Schuurmans, Dale and Bosma, Maarten and Xia, Fei and Chi, Ed and Le, Quoc V and Zhou, Denny and others},
  journal={Advances in neural information processing systems},
  volume={35},
  pages={24824--24837},
  year={2022}
}

@inproceedings{ma2025instruct,
  title={Instruct Where the Model Fails: Generative Data Augmentation via Guided Self-contrastive Fine-tuning},
  author={Ma, Weijian and Chen, Ruoxin and Zhang, Keyue and Wu, Shuang and Ding, Shouhong},
  booktitle={Proceedings of the AAAI Conference on Artificial Intelligence},
  volume={39},
  pages={5991--5999},
  year={2025}
}

@inproceedings{zheng2024llamafactory,
  title={LlamaFactory: Unified Efficient Fine-Tuning of 100+ Language Models},
  author={Yaowei Zheng and Richong Zhang and Junhao Zhang and Yanhan Ye and Zheyan Luo and Zhangchi Feng and Yongqiang Ma},
  booktitle={Proceedings of the 62nd Annual Meeting of the Association for Computational Linguistics (Volume 3: System Demonstrations)},
  address={Bangkok, Thailand},
  publisher={Association for Computational Linguistics},
  year={2024},
  url={http://arxiv.org/abs/2403.13372}
}

@article{sellergren2025medgemma,
  title={Medgemma technical report},
  author={Sellergren, Andrew and Kazemzadeh, Sahar and Jaroensri, Tiam and Kiraly, Atilla and Traverse, Madeleine and Kohlberger, Timo and Xu, Shawn and Jamil, Fayaz and Hughes, C{\'\i}an and Lau, Charles and others},
  journal={arXiv preprint arXiv:2507.05201},
  year={2025}
}

@inproceedings{pan2025medvlm,
  title={Medvlm-r1: Incentivizing medical reasoning capability of vision-language models (vlms) via reinforcement learning},
  author={Pan, Jiazhen and Liu, Che and Wu, Junde and Liu, Fenglin and Zhu, Jiayuan and Li, Hongwei Bran and Chen, Chen and Ouyang, Cheng and Rueckert, Daniel},
  booktitle={International Conference on Medical Image Computing and Computer-Assisted Intervention},
  pages={337--347},
  year={2025},
  organization={Springer}
}

@article{lai2025med,
  title={Med-r1: Reinforcement learning for generalizable medical reasoning in vision-language models},
  author={Lai, Yuxiang and Zhong, Jike and Li, Ming and Zhao, Shitian and Li, Yuheng and Psounis, Konstantinos and Yang, Xiaofeng},
  journal={arXiv preprint arXiv:2503.13939},
  year={2025}
}

@inproceedings{wu2025synthetic,
  title={Synthetic Data is an Elegant GIFT for Continual Vision-Language Models},
  author={Wu, Bin and Shi, Wuxuan and Wang, Jinqiao and Ye, Mang},
  booktitle={Proceedings of the Computer Vision and Pattern Recognition Conference},
  pages={2813--2823},
  year={2025}
}

@article{kaizu2017optical,
  title={Optical coherence tomography angiography reveals spatial bias of macular capillary dropout in diabetic retinopathy},
  author={Kaizu, Yoshihiro and Nakao, Shintaro and Yoshida, Shigeo and Hayami, Takehito and Arima, Mitsuru and Yamaguchi, Muneo and Wada, Iori and Hisatomi, Toshio and Ikeda, Yasuhiro and Ishibashi, Tatsuro and others},
  journal={Investigative Ophthalmology \& Visual Science},
  volume={58},
  number={11},
  pages={4889--4897},
  year={2017},
  publisher={The Association for Research in Vision and Ophthalmology}
}

@inproceedings{wittmann2024simulation,
  title={Simulation-based segmentation of blood vessels in cerebral 3D OCTA images},
  author={Wittmann, Bastian and Glandorf, Lukas and Paetzold, Johannes C and Amiranashvili, Tamaz and W{\"a}lchli, Thomas and Razansky, Daniel and Menze, Bjoern},
  booktitle={International Conference on Medical Image Computing and Computer-Assisted Intervention},
  pages={645--655},
  year={2024},
  organization={Springer}
}

@inproceedings{prabhakar2024vesselformer,
  title={Vesselformer: Towards complete 3d vessel graph generation from images},
  author={Prabhakar, Chinmay and Shit, Suprosanna and Paetzold, Johannes C and Ezhov, Ivan and Koner, Rajat and Li, Hongwei and Kofler, Florian Sebastian and Menze, Bjoern},
  booktitle={Medical Imaging with Deep Learning},
  pages={320--331},
  year={2024},
  organization={PMLR}
}

@article{AIREADI2024,
  author  = {{AI-READI Consortium}},
  title   = {{AI-READI}:Rethinking Data Collection, Preparation, and Sharing for Propelling AI-based Discoveries in Diabetes Research and Beyond},
  journal = {Nature Metabolism},
  year    = {2024},
  volume  = {6},
  number  = {12},
  pages   = {2210--2212},
  doi     = {10.1038/s42255-024-01165-x},
}

@inproceedings{menten2022physiology,
  title={Physiology-based simulation of the retinal vasculature enables annotation-free segmentation of OCT angiographs},
  author={Menten, Martin J and Paetzold, Johannes C and Dima, Alina and Menze, Bjoern H and Knier, Benjamin and Rueckert, Daniel},
  booktitle={International Conference on Medical Image Computing and Computer-Assisted Intervention},
  pages={330--340},
  year={2022},
  organization={Springer}
}

\newpage
\appendix

\section{Details of Vessel and Pathology Simulation}
\label{app:pathology}

This appendix provides the mathematical details of the vessel growth and pathology simulation summarized in Sec.~\ref{subsec:vessel-growth-pathology}. All coordinates are defined in a normalized 2D en-face domain $[0,1]^2$ unless otherwise stated, and radii in millimeters are mapped to pixel units via the same physical scale used by the baseline OCTA simulator~\cite{kreitner2024synthetic}.

\subsection{Baseline healthy vessel simulation}

We adopt the angiogenesis-based statistical growth model of ~\citet{kreitner2024synthetic}, which represents the vasculature as a forest of rooted binary trees growing in a 3D box
\begin{equation}
  \Omega = [0,1]\times[0,1]\times[0,h_z].
\end{equation}
Each vessel segment is an edge
\begin{equation}
  e = (\mathbf{x}_i,\mathbf{x}_j,r_{ij}),
\end{equation}
where $\mathbf{x}_i,\mathbf{x}_j\in\Omega$ are the 3D coordinates of the incident nodes and $r_{ij}>0$ is the segment radius. Sibling segments satisfy Murray’s law with bifurcation exponent $\kappa$,
\begin{equation}
  r_{\text{parent}}^{\kappa} = r_{\text{child},1}^{\kappa} + r_{\text{child},2}^{\kappa}.
\end{equation}

Growth proceeds in two phases (superficial and deep vascular complexes) by repeatedly sampling oxygen sinks (for arteries) and CO$_2$ sources (for veins) and letting leaf or inter-nodes sprout if the attraction points fall into a perception cone of distance $\delta$ and angle $\gamma$ around the current segment. The growth direction is a weighted combination of the mean attraction vector and an optimal branching vector that minimizes deviation from the parent vessel.

\paragraph{FAZ center shift}
A circular exclusion zone around the FAZ with radius $r_{\mathrm{FAZ}}$ prevents sink placement in the foveal center. To increase variability, we randomly jitter the FAZ center by a vector $\Delta\mathbf{c}$ sampled from a disk of radius $r_{\mathrm{jitter}}$ and clamp the resulting center to a maximum normalized displacement $|\Delta\mathbf{c}| \leq r_{\mathrm{max}}$. This effectively shifts the view while leaving the underlying vasculature unchanged.

After growth, the vessel graph is voxelized at high resolution and projected along the depth axis to obtain a grayscale vessel map $X$ and a binarized segmentation label $\mathrm{bin}(X)$. A pretrained GAN then converts $X$ into a realistic OCTA image.

\subsection{Capillary dropout field and vessel regression}
\label{app:dropout}

\paragraph{Dropout regions and inside-score.}
Each dropout lesion $k$ is specified by:
\begin{itemize}
  \item center $\boldsymbol{\mu}_k = (c_{k,x},c_{k,y}) \in [0,1]^2$,
  \item base radius $r_k>0$ in normalized units,
  \item axis ratios $a_k,b_k>0$ controlling ellipticity,
  \item harmonic amplitudes $A_{k,m}$ and phases $\phi_{k,m}$ for a small set of modes $\mathcal{H}_k$ (typically $m\in\{2,3,5\}$),
  \item shape exponent $\alpha_k>0$,
  \item noise gain $g_k\geq 0$ and lesion strength $s_k\in[0,1]$.
\end{itemize}

For a location $\mathbf{x}=(x,y)$ we define its coordinates relative to lesion $k$ as
\begin{equation}
  \boldsymbol{\delta}_k(\mathbf{x}) = \mathbf{x} - \boldsymbol{\mu}_k,\quad
  \rho_k(\mathbf{x}) = \|\boldsymbol{\delta}_k(\mathbf{x})\|_2,\quad
  \theta_k(\mathbf{x}) = \operatorname{atan2}(\delta_{k,y},\delta_{k,x}).
\end{equation}
The radius of the underlying ellipse in direction $\theta$ is
\begin{equation}
  R_k(\theta) =
  \frac{r_k}{\sqrt{\left(\frac{\cos\theta}{a_k}\right)^2 + \left(\frac{\sin\theta}{b_k}\right)^2}},
\end{equation}
and we modulate it with angular harmonics to obtain an irregular boundary
\begin{equation}
  \tilde{R}_k(\theta)
  =
  R_k(\theta)\left[1+\sum_{m\in\mathcal{H}_k} A_{k,m}\cos\bigl(m\theta + \phi_{k,m}\bigr)\right].
\end{equation}

We use the notation $[z]_+ = \max(z,0)$. The geometric “inside-score’’ of lesion $k$ is
\begin{equation}
  u_k(\mathbf{x}) =
  \left[
      1 - \frac{\rho_k(\mathbf{x})}{\tilde{R}_k(\theta_k(\mathbf{x}))}
  \right]_+^{\alpha_k},
\end{equation}
which smoothly decays from $1$ at the center to $0$ at the boundary.

To introduce additional boundary irregularity we construct a smooth noise field $n_k(\mathbf{x})\in[0,1]$ by summing a few sinusoidal components in $x$ and $y$. With a noise gain $g_k$ we define
\begin{equation}
  c_k(\mathbf{x}) = u_k(\mathbf{x})\,
  \mathrm{clip}\bigl(0.75 + g_k(n_k(\mathbf{x}) - 0.5),\,0,\,1.2\bigr),
\end{equation}
where $\mathrm{clip}(\cdot)$ clamps to the indicated range, and $c_k(\mathbf{x})$ is subsequently capped to $[0,1]$.

The global dropout field is given by the maximum over lesions,
\begin{equation}
  c(\mathbf{x}) = \max_k c_k(\mathbf{x})\in[0,1],
\end{equation}
and the overall dropout severity by
\begin{equation}
  s_{\max} = \max_k s_k \in [0,1].
\end{equation}

\paragraph{Probabilistic pruning of capillaries.}
Let $\mathcal{L}$ denote the set of leaf nodes (terminal segments) in one vascular forest. Each leaf $i\in\mathcal{L}$ has position $\mathbf{x}_i$ and radius $r_i$, and we evaluate $c_i = c(\mathbf{x}_i)$. We only consider leaves with
\begin{equation}
  c_i \geq \tau_{\mathrm{reg}},
\end{equation}
where $\tau_{\mathrm{reg}}\in(0,1)$ is a regression threshold (we use $\tau_{\mathrm{reg}}\approx 0.35$).

We target a global removal fraction
\begin{equation}
  f_{\mathrm{drop}} \approx s_{\max},
\end{equation}
with empirical lower/upper bounds to avoid trivial cases. Among eligible leaves we define sampling weights
\begin{equation}
  w_i \propto (1-c_i)^{\gamma}\,r_i^{-\alpha},\quad i\in\mathcal{L},
\end{equation}
where $\gamma>0$ biases removal towards the lesion core and $\alpha\in[0.3,1]$ controls the preference for smaller vessels. We sample $\lfloor f_{\mathrm{drop}}|\mathcal{L}|\rfloor$ leaves without replacement according to $w_i$ and delete the corresponding segments from the graph.

\paragraph{Remodeling: elongation and dilation.}
For non-removed nodes we model subtle remodeling. Let node $i$ with parent $p$ have positions $\mathbf{x}_i$ and $\mathbf{x}_p$, local dropout $c_i=c(\mathbf{x}_i)$, direction
\begin{equation}
  \mathbf{d}_{pi} = \mathbf{x}_i - \mathbf{x}_p,
\end{equation}
and original radius $r_i$. We sample an elongation factor $e_i\in[e_{\min},e_{\max}]$ and apply
\begin{equation}
  \mathbf{x}_i^{\mathrm{(new)}} =
  \mathbf{x}_p + \bigl[1 + (e_i-1)c_i\bigr]\,\mathbf{d}_{pi},
\end{equation}
which increases segment length more strongly near the lesion center.

Similarly, we dilate radii according to
\begin{equation}
  r_i^{\mathrm{(new)}} = r_i\left[D_{\min} + (D_{\max} - D_{\min})c_i\right],
\end{equation}
with $D_{\min},D_{\max}\geq 1$ and optional global radius clamps.

\subsection{Microaneurysm synthesis}
\label{app:ma}

Microaneurysms are modeled as short, roughly circular side branches emerging near dropout borders.

\paragraph{Spawn region.}
For each non-root arterial node $i$ with parent $p$ we compute $c_i=c(\mathbf{x}_i)$. If MAs are restricted to dropout borders, we require
\begin{equation}
  c_{\min}^{\mathrm{MA}} \leq c_i \leq c_{\max}^{\mathrm{MA}},
\end{equation}
for some band $b_{MA}:(c_{\min}^{\mathrm{MA}},c_{\max}^{\mathrm{MA}})\subset(0,1)$.

\paragraph{Spawn probability.}
For each eligible node we perform a Bernoulli trial with probability
\begin{equation}
  p_{\mathrm{MA}}(\mathbf{x}_i) =
  p_0\bigl(1+\lambda_s s_{\max}\bigr)\bigl(1+\lambda_c c_i\bigr),
\end{equation}
where $p_0$ is a base MA density, $\lambda_s$ couples MA counts to global dropout severity, and $\lambda_c$ increases density near dropout borders. We also reweight $p_{\mathrm{MA}}$ by the area of the largest dropout region, so larger lesions tend to host more MAs.

\paragraph{MA geometry.}
We form a short side branch. Let
\begin{equation}
  \mathbf{u}_{pi}=\frac{\mathbf{x}_i-\mathbf{x}_p}{\|\mathbf{x}_i-\mathbf{x}_p\|_2}
\end{equation}
be the parent-to-child direction and
\begin{equation}
  \mathbf{u}_{\perp}=(-u_{pi,y},u_{pi,x},0)
\end{equation}
a perpendicular unit vector in the en-face plane. With simulator step size $d$ and MA length factor $\ell_{\mathrm{MA}}$ we place the MA center at
\begin{equation}
  \mathbf{x}_{\mathrm{MA}} = \mathbf{x}_i + \ell_{\mathrm{MA}}d\,\mathbf{u}_{\perp}.
\end{equation}
The MA radius is sampled as
\begin{equation}
  r_{\mathrm{MA}} \sim \mathcal{U}\bigl(r_{\min}^{\mathrm{MA}},r_{\max}^{\mathrm{MA}}\bigr),
\end{equation}
where $r_{\min}^{\mathrm{MA}}$ and $r_{\max}^{\mathrm{MA}}$ are specified in millimeters. Additional child nodes can be sampled in a small disk around $\mathbf{x}_{\mathrm{MA}}$ with slightly reduced radii to create irregular clusters.

\subsection{Neovascular tufts}
\label{app:nv}

Neovascularization is represented as thin, tortuous sprouts grown from existing vessel tips.

\paragraph{Tip selection.}
We project arterial segments to 2D, identify leaf tips $j$ with position $\mathbf{x}_j$, tangent direction $\mathbf{u}_j$, and local dropout value $c(\mathbf{x}_j)$. Tips too close to the FAZ or clearly outside dropout are down-weighted. A global NV severity parameter $s_{\mathrm{NV}}\in[0,1]$ determines the number of NV groups $G$ and typical sprout length.

\paragraph{Main tuft growth.}
For each selected tip we initialize a polyline $\{\mathbf{p}^{(g)}_t\}_{t=0}^{T_{\mathrm{main}}}$ with
\begin{equation}
  \mathbf{p}^{(g)}_0 = \mathbf{x}_j.
\end{equation}
At iteration $t$ we update
\begin{equation}
  \mathbf{p}^{(g)}_{t+1}
  =
  \mathbf{p}^{(g)}_t + \ell_{\mathrm{NV}}\mathbf{v}^{(g)}_t,
\end{equation}
where $\ell_{\mathrm{NV}}$ is a small step size and $\mathbf{v}^{(g)}_t$ is a unit direction obtained as a weighted combination of:
\begin{enumerate}
  \item previous direction $\mathbf{v}^{(g)}_{t-1}$ (persistence),
  \item a weak radial vector pointing away from the closest dropout center,
  \item a low-frequency swirling field and isotropic jitter.
\end{enumerate}
We clamp positions to remain within dropout regions and outside the FAZ.

\paragraph{Radius profile and side branches.}
Along each polyline we use a linearly tapering radius
\begin{equation}
  r_{\mathrm{NV}}(t) =
  (1-\tau_t)r_{\mathrm{start}} + \tau_t r_{\mathrm{end}},\quad
  \tau_t=\frac{t}{T_{\mathrm{main}}},
\end{equation}
with $r_{\mathrm{start}}$ proportional to the parent vessel radius and $r_{\mathrm{end}}<r_{\mathrm{start}}$. With probability increasing in $s_{\mathrm{NV}}$ we spawn side branches starting from intermediate points. These follow the same growth rule but with shorter maximum length $T_{\mathrm{side}}$.

\subsection{Tortuosity along dropout borders}
\label{app:tortuosity}

To model increased tortuosity at dropout borders we jitter node positions perpendicular to the local vessel direction within a band of the dropout field.

For a non-root arterial node $i$ with parent $p$, we define the tangent direction
\begin{equation}
  \mathbf{u}_{pi} = \frac{\mathbf{x}_i-\mathbf{x}_p}{\|\mathbf{x}_i-\mathbf{x}_p\|_2},
\end{equation}
and choose a perpendicular direction $\mathbf{u}_{\perp}$ as above. Let $c_i=c(\mathbf{x}_i)$ and specify a tortuosity band $b_{tort}$ where
\begin{equation}
  c_{\min}^{\mathrm{tort}} \leq c_i \leq c_{\max}^{\mathrm{tort}}.
\end{equation}
Nodes outside this band are left unchanged.

Given a gain parameter $g_{\mathrm{tort}}\in[0,1]$ and simulator step size $d$, we set the jitter amplitude
\begin{equation}
  A_{\mathrm{tort}} = 0.35\,g_{\mathrm{tort}}\,d,
\end{equation}
sample
\begin{equation}
  \epsilon_i \sim \mathcal{U}(-A_{\mathrm{tort}},A_{\mathrm{tort}}),
\end{equation}
and update the node position as
\begin{equation}
  \mathbf{x}_i^{\mathrm{(new)}} = \mathbf{x}_i + \epsilon_i\,\mathbf{u}_{\perp},
\end{equation}
clipping back to $[0,1]^2$ if necessary. This leaves the topology and global FAZ shape intact but increases local curvature.

\subsection{Parameter ranges and typical settings}
\label{app:param-ranges}

Tab.~\ref{tab:pathology-params} summarizes the key parameters that control the four DR lesion types and the typical ranges used in the OCTA-100K-SVR dataset. Spatial coordinates are normalized to $[0,1]^2$; radii in millimeters refer to the physical scale of the baseline simulator.

\begin{table}[H]
  \centering
  \caption{Key parameters of the simulator and typical settings used in this work.}
  \label{tab:pathology-params}
  \small
  \setlength{\tabcolsep}{4pt}
  \begin{tabular}{lll}
    \toprule
    Pathology & Parameter & Typical value / range \\
    \midrule
    \multirow{5}{*}{Dropout}
      & \# regions $n$      & $[0, 6]$ per sample \\
      & Radius $r_{\mathrm{drop}}$          & $[0.18, 0.32]$ (normalized) \\
      & Lesion strength $s_k$              & $[0.90, 0.99]$ \\
      & Gradient exponent $\alpha_k$       & $[2.0, 3.0]$ \\
      & Noise gain $g_k$                   & $[0.20, 0.40]$ \\
    \midrule
    \multirow{4}{*}{MA}
      & Base density $p_0$                 & $\approx 0.03$ (per segment) \\
      & Radius $r_{\mathrm{MA}}$           & $[0.01, 0.08]\,\mathrm{mm}$ \\
      & Length factor $\ell_{\mathrm{MA}}$ & $[0.3, 0.4]$ (in units of $d$) \\
      & Strength coupling $\lambda_s$      & $\approx 15$ \\
    \midrule
    \multirow{4}{*}{NV}
      & NV probability                     & $\approx 0.4$ \\
      & Severity $s_{\mathrm{NV}}$         & $[0.2, 0.7]$ \\
      & Footprint radius                   & $[0.015, 0.07]$ (normalized) \\
      & Sprout length $l_{\mathrm{NV}}$    & $[3, 6]$ growth steps \\
    \midrule
    \multirow{2}{*}{Tortuosity}
      & Gain $g_{\mathrm{tort}}$           & $[0.01, 0.5]$ \\
      & Band $[c_{\min}^{\mathrm{tort}}, c_{\max}^{\mathrm{tort}}]$
                                          & $[0.30, 0.75]$ \\
    \bottomrule
  \end{tabular}
\end{table}

\subsection{Extra Examples of Reasoning Texts}
\label{app:reasoning_egs}

\begin{figure}[H]
    \centering
    \includegraphics[width=1.0\linewidth]{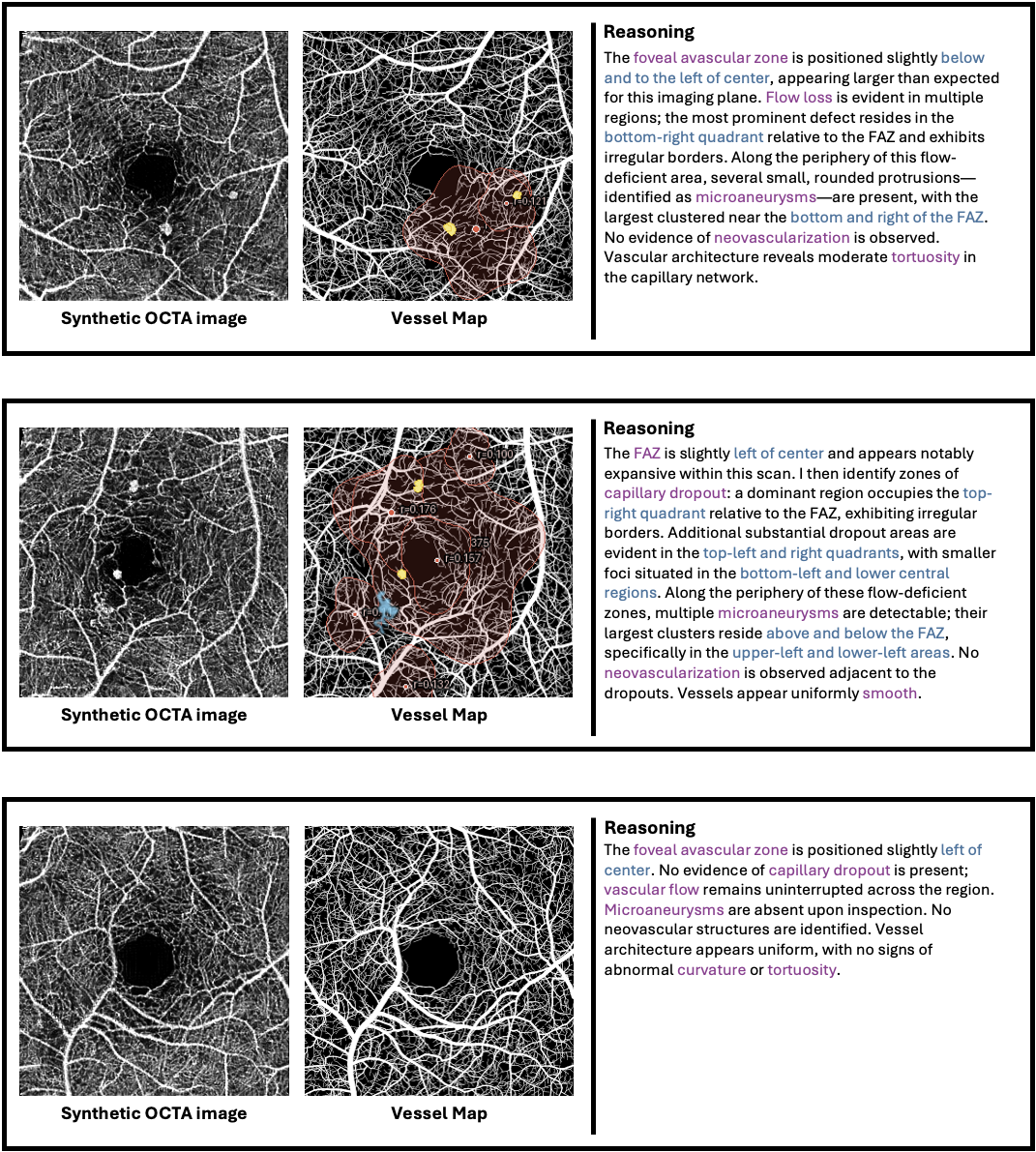}
    \caption{Extra Examples of Synthetic OCTA and Reasoning.}
    \label{fig:reasoning_2}
\end{figure}

\newpage

\section{Pseudocode of the SVR Pipeline}
\label{app:pseudocode}

\RestyleAlgo{plain}  

\begin{mdframed}[roundcorner=3pt, linewidth=0.6pt]
\begin{algorithm}[H]
\DontPrintSemicolon
\SetAlgoLined
\caption{Synthetic Vasculature Reasoning (SVR) pipeline}
\label{alg:svr-pipeline}
\KwIn{number of synthetic samples $N$; vessel-growth config $\theta_{\mathrm{vessel}}$;
      pathology profile $\theta_{\mathrm{path}}$; optional GAN config $\theta_{\mathrm{GAN}}$;
      teacher VLM $T$; base VLM $M$}
\KwOut{synthetic image–text dataset $\mathcal{D}$; pretrained model $M_{\mathrm{SVR}}$; fine-tuned model $M_{\mathrm{SVR\text{-}FT}}$}

Initialize dataset $\mathcal{D} \leftarrow \varnothing$\;
\For{$i = 1$ \KwTo $N$}{
  Sample structural parameters (FAZ, seed, view shift, etc.)\;
  
  Simulate healthy vessel graph $G_i$
  
  Apply pathology profile $\theta_{\mathrm{path}}$ to get augmented graph $G_i^\ast$\; 
  
  Rasterize $G_i^\ast$ into a 2D vessel map $X_i$\; 
  
  \eIf{GAN is enabled}{
    Generate OCTA-like image $I_i$ from $X_i$ using $\theta_{\mathrm{GAN}}$\;
  }{
    Set $I_i \leftarrow X_i$\;
  }

  Extract structured metadata $m_i$ from the simulator (FAZ, dropout, MA, NV, tortuosity)\; 
  
  Build template reasoning $r_i^{\mathrm{tmp}}$ from $m_i$ (FAZ $\rightarrow$ dropout $\rightarrow$ MA $\rightarrow$ NV $\rightarrow$ tortuosity)\;
  
  Query teacher VLM $T$ with $(I_i, m_i, r_i^{\mathrm{tmp}})$ to obtain a diversified, fact-preserving reasoning $r_i$\;

  Define question $q_i$ asking to describe $I_i$ and assess DR stage\;
  
  Define answer $a_i$ as $r_i$ (plus final diagnosis sentence for clinical data)\;
  Add sample $(I_i, q_i, a_i, m_i)$ to dataset $\mathcal{D}$\;
}

Pretrain base VLM $M$ on $\mathcal{D}$ (freeze language backbone) to obtain $M_{\mathrm{SVR}}$\;
Fine-tune $M_{\mathrm{SVR}}$ end-to-end on clinical OCTA data with reasoning and DR labels to obtain $M_{\mathrm{SVR\text{-}FT}}$\;
\end{algorithm}
\end{mdframed}

\section{Training details}
\label{app:training}

\paragraph{SVR training prompt (student VLM).}
During both SVR pretraining and SVR-FT fine-tuning, the VLM is trained in a single-turn instruction-following format. The human turn consists of the OCTA image token plus the following question (identical for synthetic and clinical data):
\begin{verbatim}
<image>
What features are visible in this OCTA image? 
Please first describe the image features, then inspect it for signs 
of diabetic retinopathy (DR) and classify it as Healthy, NPDR, or PDR.
\end{verbatim}
The assistant turn contains the full Chain-of-Thought reasoning followed by the final diagnosis sentence.

\paragraph{CoT diversification prompts (teacher VLM).}
To obtain linguistically diverse and fact-preserving reasoning, we prompt GPT-5 with a system prompt and a user prompt that include the synthetic OCTA image, the original CoT and structured pathology metadata. The system prompt is:
\begin{verbatim}
You are an ophthalmology OCTA expert and skilled medical writer.
You will receive an OCTA image, concise metadata, and an original
chain-of-thought (CoT). Rewrite the CoT in a different language
style while preserving ALL medical facts, locations, and uncertainty.
Do not add new findings. Keep content consistent with the image and
metadata. Spatial terminology constraint: avoid eye-dependent terms
(e.g., temporal, nasal, superotemporal, inferonasal, superior/inferior
when tied to eye laterality). Use only absolute image directions such
as left, right, up, down, and center to describe locations.
Aim for similar length and clarity. Output only the rewritten CoT.
\end{verbatim}
The corresponding user prompt is:
\begin{verbatim}
Here is an OCTA image <image>.
Metadata (JSON): <COMPACT_METADATA_JSON>

Original CoT describing the image:
<ORIGINAL_COT>

Task: Rewrite the CoT with a distinct language style (e.g., more
academic, more succinct, or slightly conversational) while preserving
all facts and spatial relations. Do not invent new content. 
Use only absolute image directions such as left, right, up, down, and center.
Return only the rewritten CoT.
\end{verbatim}
where \verb|<COMPACT_METADATA_JSON>| is the compacted pathology metadata (FAZ, dropout, MA, NV, tortuosity) extracted from the simulator and \verb|<ORIGINAL_COT>| is the deterministic template reasoning.

For robustness to natural-language variation in the input question, we also diversify the user prompt itself using a similar pair of prompts.

\section{Supplementary Details in Evaluation}
\label{app:evaluation-details}

At inference we use temperature $0.1$ and top-$p=0.8$, and compute all metrics from unified JSONL outputs to ensure consistent parsing across models. To evaluate using GPT-5 scores, for each image, question, and model response, GPT-5 receives (i) the OCTA image, (ii) the dataset tag (synthetic vs.\ in-house), (iii) the question text, (iv) the ground-truth DR label when available, (v) an optional synthetic reference explanation (for synthetic data), and (vi) the candidate response. The system message is the fixed instruction:
\begin{verbatim}
You are a retina specialist. First write your response to the question with 
all the information provided, and then score a single model response 
by comparing it with yours on:
- helpfulness (0-10): clarity, usefulness, specificity to the question.
- accuracy (0-10): localization + clinical correctness (use the image;
  if GT label provided, consider it).
- relevance (0-10): how on-topic and focused the response is.
Return a compact JSON: {"helpfulness": int, "accuracy": int,
"relevance": int, "rationale": string}.
\end{verbatim}
and the user message concatenates the textual context (dataset, question, GT label, reference, model response) with the OCTA image. We then parse the \texttt{helpfulness}, \texttt{accuracy}, and \texttt{relevance} integers and average them to obtain the per-sample ``GPT-5 Score'', which is then aggregated over all samples for each model and dataset.

\begin{table}[H]
\centering
\caption{
DR staging on the in-house OCTA dataset.
H = Healthy, N = NPDR, P = PDR.
}
\label{tab:inhouse}
\resizebox{0.98\columnwidth}{!}{
\begin{tabular}{lccccccc}
\toprule
\textbf{Model}
& \textbf{Prec(H)} & \textbf{Prec(N)} & \textbf{Prec(P)}
& \textbf{Rec(H)} & \textbf{Rec(N)} & \textbf{Rec(P)}
& \cellcolor{gray!15}{\textbf{Bal.Acc}} \\
\midrule

ResNet18~\cite{lux2025interpretable}
& \begin{tabular}[t]{@{}c@{}}0.943 \\ {\scriptsize$\pm$\,0.008}\end{tabular}
& \begin{tabular}[t]{@{}c@{}}0.335 \\ {\scriptsize$\pm$\,0.044}\end{tabular}
& \begin{tabular}[t]{@{}c@{}}0.426 \\ {\scriptsize$\pm$\,0.070}\end{tabular}
& \begin{tabular}[t]{@{}c@{}}0.793 \\ {\scriptsize$\pm$\,0.021}\end{tabular}
& \begin{tabular}[t]{@{}c@{}}0.544 \\ {\scriptsize$\pm$\,0.094}\end{tabular}
& \begin{tabular}[t]{@{}c@{}}0.563 \\ {\scriptsize$\pm$\,0.131}\end{tabular}
& \cellcolor{gray!15}{\begin{tabular}[t]{@{}c@{}}0.633 \\ {\scriptsize$\pm$\,0.030}\end{tabular}} \\

GNN~\cite{lux2025interpretable}
& \begin{tabular}[t]{@{}c@{}}0.950 \\ {\scriptsize$\pm$\,0.010}\end{tabular}
& \begin{tabular}[t]{@{}c@{}}0.326 \\ {\scriptsize$\pm$\,0.049}\end{tabular}
& \begin{tabular}[t]{@{}c@{}}\textbf{0.456} \\ {\scriptsize$\pm$\,0.107}\end{tabular}
& \begin{tabular}[t]{@{}c@{}}0.720 \\ {\scriptsize$\pm$\,0.035}\end{tabular}
& \begin{tabular}[t]{@{}c@{}}0.594 \\ {\scriptsize$\pm$\,0.091}\end{tabular}
& \begin{tabular}[t]{@{}c@{}}\textbf{0.775} \\ {\scriptsize$\pm$\,0.034}\end{tabular}
& \cellcolor{gray!15}{\begin{tabular}[t]{@{}c@{}}\textbf{0.697} \\ {\scriptsize$\pm$\,0.034}\end{tabular}} \\

Qwen3-VL-8b~\cite{yang2025qwen3}
& 0.775 & 0.198 & 0.065
& 0.576 & 0.163 & 0.255 & \cellcolor{gray!15}{0.331} \\

Qwen3-VL-30b~\cite{yang2025qwen3}
& 0.787 & 0.228 & 0.444
& \textbf{0.849} & 0.241 & 0.041 & \cellcolor{gray!15}{0.377} \\

Qwen3-VL-8b-GFT~\cite{li2025fine}
& \begin{tabular}[t]{@{}c@{}}0.889 \\ {\scriptsize$\pm$\,0.007}\end{tabular}
& \begin{tabular}[t]{@{}c@{}}\textbf{0.535} \\ {\scriptsize$\pm$\,0.045}\end{tabular}
& \begin{tabular}[t]{@{}c@{}}0.454 \\ {\scriptsize$\pm$\,0.017}\end{tabular}
& \begin{tabular}[t]{@{}c@{}}0.881 \\ {\scriptsize$\pm$\,0.019}\end{tabular}
& \begin{tabular}[t]{@{}c@{}}0.502 \\ {\scriptsize$\pm$\,0.027}\end{tabular}
& \begin{tabular}[t]{@{}c@{}}0.550 \\ {\scriptsize$\pm$\,0.041}\end{tabular}
& \cellcolor{gray!15}{\begin{tabular}[t]{@{}c@{}}0.645 \\ {\scriptsize$\pm$\,0.017}\end{tabular}} \\

Qwen3-VL-8b-FT
& \begin{tabular}[t]{@{}c@{}}0.908 \\ {\scriptsize$\pm$\,0.014}\end{tabular}
& \begin{tabular}[t]{@{}c@{}}0.364 \\ {\scriptsize$\pm$\,0.084}\end{tabular}
& \begin{tabular}[t]{@{}c@{}}0.245 \\ {\scriptsize$\pm$\,0.322}\end{tabular}
& \begin{tabular}[t]{@{}c@{}}0.828 \\ {\scriptsize$\pm$\,0.071}\end{tabular}
& \begin{tabular}[t]{@{}c@{}}\textbf{0.636} \\ {\scriptsize$\pm$\,0.061}\end{tabular}
& \begin{tabular}[t]{@{}c@{}}0.073 \\ {\scriptsize$\pm$\,0.107}\end{tabular}
& \cellcolor{gray!15}{\begin{tabular}[t]{@{}c@{}}0.512 \\ {\scriptsize$\pm$\,0.050}\end{tabular}} \\

Qwen3-VL-8b-SVR (ours)
& 0.928 & 0.295 & 0.357
& 0.786 & 0.453 & 0.510 & \cellcolor{gray!15}{0.583} \\

Qwen3-VL-8b-SVR-FT (ours)
& \begin{tabular}[t]{@{}c@{}}\textbf{0.958} \\ {\scriptsize$\pm$\,0.024}\end{tabular}
& \begin{tabular}[t]{@{}c@{}}0.411 \\ {\scriptsize$\pm$\,0.087}\end{tabular}
& \begin{tabular}[t]{@{}c@{}}0.418 \\ {\scriptsize$\pm$\,0.016}\end{tabular}
& \begin{tabular}[t]{@{}c@{}}0.840 \\ {\scriptsize$\pm$\,0.082}\end{tabular}
& \begin{tabular}[t]{@{}c@{}}0.488 \\ {\scriptsize$\pm$\,0.088}\end{tabular}
& \begin{tabular}[t]{@{}c@{}}0.737 \\ {\scriptsize$\pm$\,0.074}\end{tabular}
& \cellcolor{gray!15}{\begin{tabular}[t]{@{}c@{}}0.688 \\ {\scriptsize$\pm$\,0.027}\end{tabular}} \\

\bottomrule
\end{tabular}}
\end{table}

\begin{figure}[ht]
    \centering
    \includegraphics[width=0.7\linewidth]{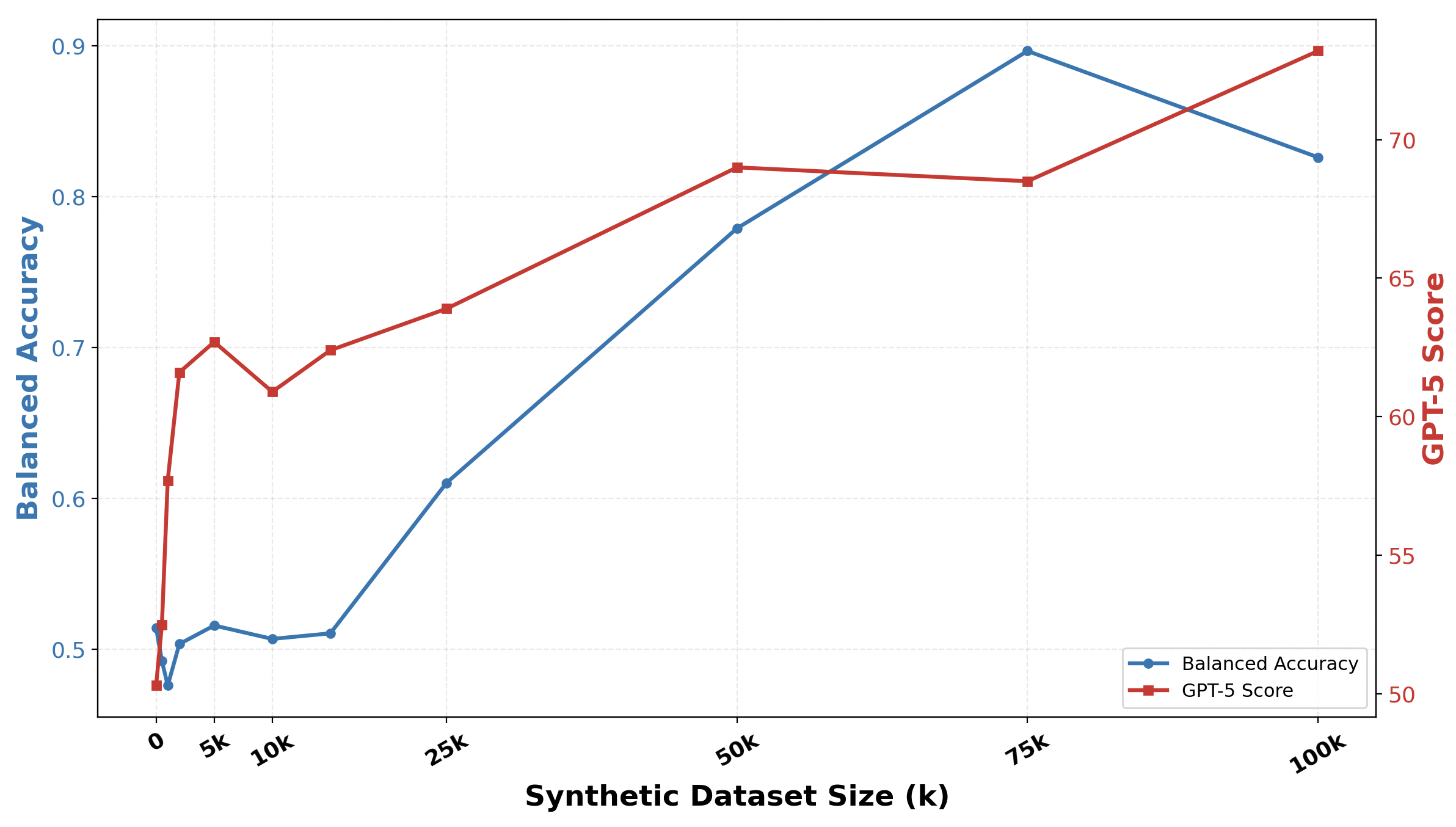}
    \caption{Correlation between Classification Performance and GPT-5 Score}
    \label{fig:correlation}
\end{figure}



\begin{figure}[H]
    \centering
    \includegraphics[width=0.9\linewidth]{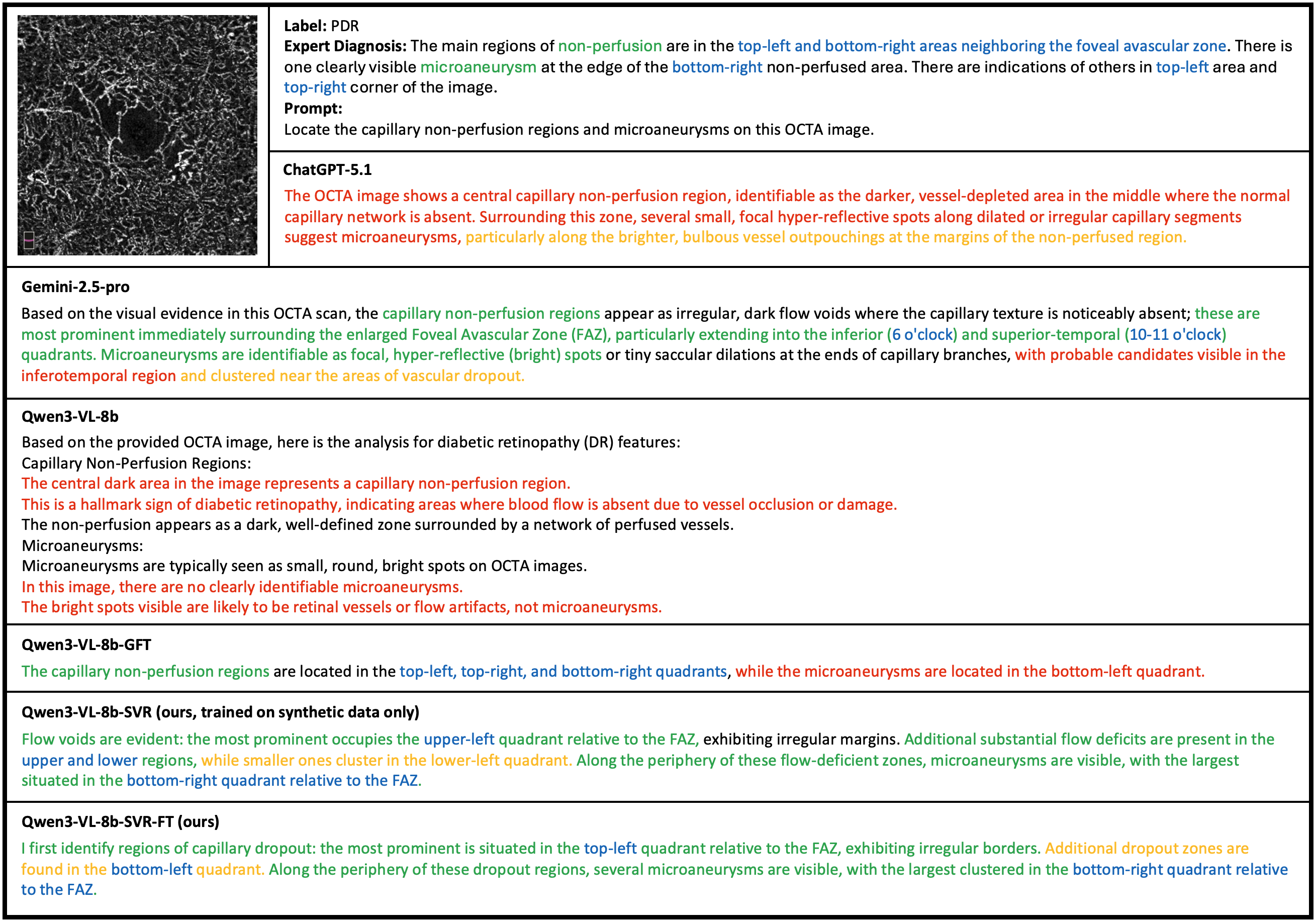}
    \caption{Extra Example for Qualitative Evaluation.}
\end{figure}

\begin{figure}[H]
    \centering
    \includegraphics[width=0.9\linewidth]{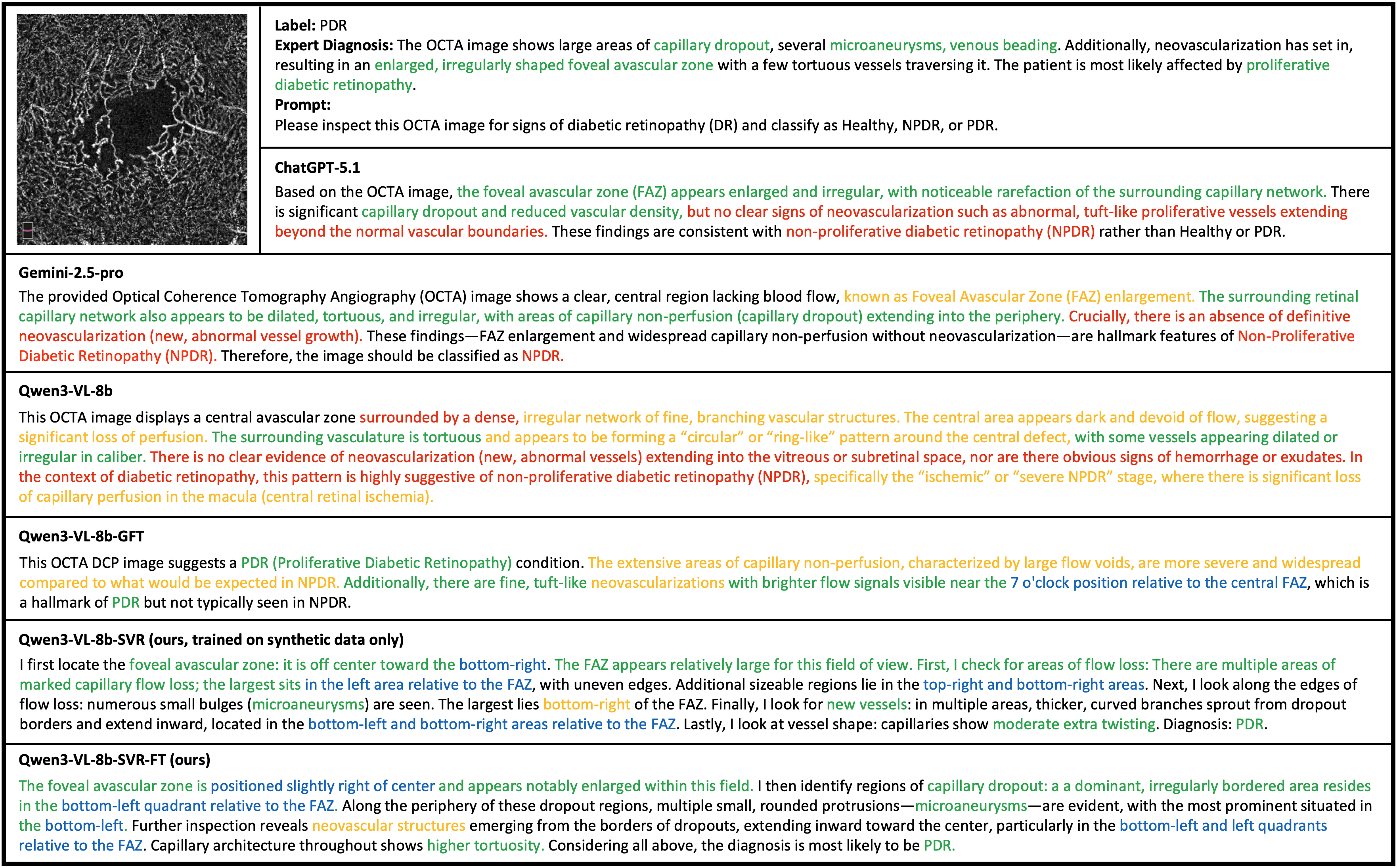}
    \caption{Extra Example for Qualitative Evaluation.}
\end{figure}

\end{document}